\documentclass[sigconf,10pt]{acmart}
\AtBeginDocument{%
  }

\renewcommand\footnotetextcopyrightpermission[1]{}
\setcopyright{none}
\copyrightyear{}

\usepackage{graphicx}
\usepackage[inline]{enumitem}
\usepackage{subcaption}
\usepackage{multirow}
\usepackage[table,xcdraw]{xcolor}
\usepackage{xr}
\usepackage{colortbl}
\usepackage{xcolor}
\usepackage{tikz}
\usetikzlibrary{positioning,arrows.meta}

\usepackage{microtype}
\usepackage[T1]{fontenc} 
\usepackage{tgtermes}

\usepackage{cleveref}
\crefformat{figure}{#2#1#3} 
\crefrangeformat{figure}{#3#1#4--#5#2#6} 
\crefmultiformat{figure}{#2#1#3}{ and #2#1#3}{, #2#1#3}{, and #2#1#3}  
\crefformat{table}{#2#1#3} 
\crefrangeformat{table}{#3#1#4--#5#2#6} 
\crefmultiformat{table}{#2#1#3}{ and #2#1#3}{, #2#1#3}{, and #2#1#3}  

\acmDOI{}
\acmConference[]{}{}{}
\acmISBN{}

\newcommand{\br}[1]{\textcolor{red}{[Bozidar: #1]}}

\newcommand{\gaa}[1]{\textcolor{orange}{[#1]}}

\newcommand{\xref}[1]{§\ref{#1}}

\begin{document}

\title{Offload or Overload: A Platform Measurement Study of Mobile Robotic Manipulation Workloads}

\author[Sara Pohland et al.]{   
Sara Pohland$^{1}$ 
Xenofon Foukas$^{2}$, 
Ganesh Ananthanarayanan$^{2}$, 
Andrey Kolobov$^{2}$, 
Sanjeev Mehrotra$^{2}$, 
Bozidar Radunovic$^{2}$, 
Ankit Verma$^{2}$\\
$^{1}$UC Berkeley, $^{2}$Microsoft Corp.}

\begin{abstract}
Mobile robotic manipulation--the ability of robots to navigate spaces and interact with objects--is a core capability of physical AI. 
Foundation models have led to breakthroughs in their performance, but at a significant computational cost. 
We present the first measurement study of mobile robotic manipulation workloads across onboard, edge, and cloud GPU platforms. 
We find that the full workload stack is infeasible to run on smaller onboard GPUs, while larger onboard GPUs drain robot batteries several hours faster. Offloading alleviates these constraints but introduces its own challenges, as additional network latency degrades task accuracy, and the bandwidth requirement makes naive cloud offloading impractical. Finally, we quantify opportunities and pitfalls of sharing compute across robot fleets.
We believe our measurement study will be crucial to designing inference systems for mobile robots.
\end{abstract}

\maketitle
\pagestyle{plain}

\section{Introduction}
\label{sec:intro}



While AI-based systems, such as ChatGPT and Claude, have transformed digital workflows, they remain largely disconnected from the physical world, leaving much of AI’s potential untapped. We believe the next wave of AI will center on \textit{physical AI}: embodied robotic assistants deployed as coordinated fleets in homes, enterprises, and industrial settings. Companies like Tesla, Google, and Figure AI are already pursuing general physical intelligence, signaling an industry-wide push.

A core capability underlying physical AI is \textit{mobile robotic manipulation}--the ability of robots to navigate spaces and interact with objects. This capability is the building block for multi-step tasks that combine navigation (moving to a target) with object manipulation (moving objects). Mobile manipulation unlocks real-world applications across homes (e.g., clearing a countertop), enterprises (e.g., delivering mail), and industrial settings (e.g., moving tools). Recent advances in foundation models have driven breakthroughs in mobile robotic manipulation, enabling generalization across tasks, objects, environments, and robotic embodiments.


However, these advances come at a {\em high computational cost}. Robotics foundation models are orders of magnitude larger than the classical models and expert system policies traditionally used in robotics. Modern robots can be equipped with powerful onboard CPUs and GPUs, but these compute platforms can lead to a significant increase in the robot's cost and a decrease in its battery lifetime. While lightweight onboard GPUs are an option, they present a hard tradeoff by increasing the execution time of the foundation models, thus reducing the robot's practical utility. Lighter models can reduce the execution time but at the expense of accuracy.


\noindent{\bf Objectives:} We present the {\em first measurement study of workloads involved in mobile robotic manipulation} on onboard, edge, and cloud GPU platforms, with both wireless and wired connectivity, to address the following key questions:
\begin{itemize}[leftmargin=*]
\item How severe are the tradeoffs between memory usage, task performance, and power consumption for onboard GPUs?
\item If workloads were to be offloaded, what are the effects of the added network constraints (latency and bandwidth) on the execution time and accuracy of the workloads?
\item When offloading, what are the implications of having multiple robots share the network and GPU compute? 
\end{itemize}
We argue that answering these questions will enable the design of more effective mobile robotic manipulation systems.


Our measurement study considers different robot form factors, workloads for mobile robotic manipulation, GPU computes and wireless connectivity. The SO-101, TurtleBot 4, and Stretch 3 robots represent distinct points in the design space of robots in cost and capabilities; \xref{sec:setup}. We consider a wide variety of GPUs for onboard compute (Jetson Orin and Thor), edge compute (L4 and DGX Spark), and cloud compute (A100 GPU), as well as wireless connectivity for the robots; 
\xref{sec:setup}. Finally, we use state-of-the-art robotic workloads for perception, planning, navigation, and manipulation; \xref{sec:methodology}. Our extensive setup gives us confidence that our results can broadly inform decisions in designing robotic inference systems.

{\bf 1) Onboard GPU robotics:} With onboard robot GPUs, we observe that smaller GPUs, such as the Jetson Orin with 32GB of memory, cannot even fit the full stack of models for mobile robotic manipulation. On GPUs with sufficient memory, mapping and planning can be slower by up to 383\% compared to an A100, thus limiting the robot's abilities in dynamic spaces. Navigation sees a 30\% drop in its timely detection of obstacles with lighter GPUs. While the manipulation VLA models do not dramatically slow down with smaller GPUs, the slowdown is still sufficient to drop their accuracies by 50\%. Finally, the larger onboard GPUs such as Jetson Thor drain robot batteries considerably faster, by up to 160\% (or a few hours), for even the larger robots. Our study in \xref{sec:on_board} concludes that the impact on accuracy, execution time, and battery lifetime of onboard GPUs offers a case for offloading model inference.

{\bf 2) Offloading considerations:} Offloading robotics AI inference, nonetheless, presents a non-trivial design space. We quantify that inference times on cloud-class GPUs are considerably smaller than on edge-class GPUs, but the network connectivity to reach the GPU for offloading presents a challenge. Even a few tens of milliseconds of additional latency can drop the accuracy of manipulation by over 10\%. Further, offloading requires continuously transmitting image streams from the robot, which can saturate the network. Attempts to compress the streams can drop accuracy of manipulation and semantic mapping by nearly 20\%. Thus, offloading is not a panacea and requires straddling a complex tradeoff of GPUs, connectivity, and video compression, as we explain in \xref{sec:offloading}.

{\bf 3) Multiplexing robotic offloads:} Offloading inference of a {\em fleet} of robots to an edge compute (or cloud) presents both challenges and opportunities. We show that offloading compute to a shared platform provides opportunity to batch requests across multiple robots and to leverage statistical multiplexing across time-varying workloads. This means that we can use less compute to serve the same workload.  
However, the inference time and network latencies rise under contention from multiple robots, thus an offloading solution requires the use of admission control or careful resource management with performance guarantees for high-priority robotics tasks.

In summary, we make the following contributions: $(i)$ We identify the emerging importance of mobile robotics manipulation and design a suite of representative workloads for benchmarking. $(ii)$ We analyze the complex tradeoff in latency, accuracy, and power between equipping robots with onboard GPUs and offloading their inference to the edge or cloud. $(iii)$ We present the opportunities and challenges in compute and communication with multiplexing inference of large robot fleets.
We believe that our extensive study will help guide further work on inference systems for mobile robots.

\section{Background \& Motivation}
\label{sec:background}



In this section, we describe the recent work in robotics and compute infrastructure that motivated our studies.

\subsection{Mobile robotic manipulation}
\label{subsec:overview}




Mobile manipulation is a fundamental capability in robotics that promises to significantly alter the future of robotics~\cite{colossus_robots_mainstream}, with potential impacts across important areas of society, ranging from reindustrialization~\cite{semianalysis_labor_robotics_1} to household assistance and elderly care~\cite{bbc_robotics_2024}.
Recent advances in foundation models are beginning to enable mobile manipulation to move beyond the highly structured factory environments into less structured, open-world environments~\cite{semianalysis_robotics_autonomy_levels, colossus_robots_mainstream}.
Mobile robotic manipulation integrates four key competencies: perception, planning, navigation, and manipulation~\cite{ovmm} (summarized in Table \ref{tab:applications-table}).


\begin{table}[t]
\centering
\footnotesize
\begin{tabular}{|c|c|c|}
\hline
\textbf{Task type} & \textbf{Focus} & \textbf{Examples} \\ \hline

Perception                                                         
& \begin{tabular}[c]{@{}c@{}}Sense \& understand \\ the environment\end{tabular}  & \begin{tabular}[c]{@{}c@{}}SLAM, object detection, \\ semantic mapping \end{tabular} 
\\ \hline

Planning 
& \begin{tabular}[c]{@{}c@{}} Interpret high-level \\ goals \& plan actions\end{tabular} &
\begin{tabular}[c]{@{}c@{}} Command interpretation, \\ task sequencing \end{tabular}
\\ \hline

Navigation
& \begin{tabular}[c]{@{}c@{}} Move safely \& efficiently \\ to a target location \end{tabular} & \begin{tabular}[c]{@{}c@{}} Path following, \\ obstacle avoidance \end{tabular}
\\ \hline
Manipulation                                                       & \begin{tabular}[c]{@{}c@{}} Interact physically \\ with objects  \end{tabular} & \begin{tabular}[c]{@{}c@{}} Grasping, lifting, \\ opening, placing \end{tabular}
\\ \hline

\end{tabular}

\caption{\small Examples of typical tasks required to enable a mobile robotic manipulation application.}
\label{tab:applications-table}
\vspace{-1cm}
\end{table}

\textbf{1) Perception}--i.e., using onboard sensors to collect data and generate a representation of the robot's surroundings--is tightly intertwined with the other competencies and often involves simultaneous localization and mapping (SLAM), whereby the robot tries to estimate its own location while constructing a map of the environment~\cite{slam}. In contrast to geometric grid maps, robots increasingly integrate open-vocabulary object detection and segmentation~\cite{perception_survey} to obtain semantically rich representations for detailed reasoning~\cite{mapping_survey}.  

\textbf{2)} After perceiving its environment, the robot generates a \textbf{plan} to accomplish its assigned task, 
which involves task decomposition. For example, to collect all dirty dishes, the robot must search the office for dish-like objects, then navigate to each dish and grasp it, and finally transport the collected dishes to the kitchen sink. 
In mobile manipulation, planning involves grounding natural language instructions into actions based on the robot's environmental perception
~\cite{alfred,language_grounding}.

\textbf{3)} The steps planned by the robot typically involve navigation and manipulation. In \textbf{navigation}, the goal is to move from one location to another while avoiding collisions with obstacles and other agents and possibly following a low-cost or socially compliant path~\cite{human_nav}. Navigation requires both global planning on a map~\cite{path_planning} and local planning to react to dynamic or unforeseen obstacles~\cite{local_control}.

\textbf{4)} Finally, \textbf{manipulation} is the physical interaction with objects via, e.g., grasping, lifting, opening, or placing~\cite{manipulation}. Successful manipulation requires implicit or explicit comprehension of geometry, pose uncertainty, contact dynamics, and task-specific constraints, all of which become substantially more challenging in unstructured human environments~\cite{limitations-1}, which is the focus of recent work~\cite{manipulation-2}.

\subsection{Advances in robotics capabilities}
\label{subsec:foundation}

With the advent of \textit{foundation models} containing billions of parameters, the ability of robots to engage in mobile manipulation has dramatically changed. Combining large language models (LLMs) (e.g., GPT~\cite{gpt4}) with vision-language embedding models (e.g., CLIP~\cite{clip}) has led to vision language models (VLMs) such as LLaVA \cite{llava} and GPT-4V \cite{gpt4}. VLMs provide the backbone for open-vocabulary object detection models, semantic segmentation models (e.g., SAM~\cite{sam}), 3D world modeling systems (e.g., VLMaps~\cite{vlmaps}), vision language action (VLA) models for robot manipulation (e.g., $\pi_{0.5}$~\cite{pi0.5}),
and planning capability for embodied question answering (EQA) (e.g., GraphEQA \cite{grapheqa}). Additionally, the emergence of video models has led to the development of World Action Models (WAMs) (e.g., DreamZero~\cite{ye2026world}).
Alongside advancements in foundation models, improvements in SLAM have made perception and navigation more robust in diverse and dynamic environments, while reducing navigation times, avoiding expensive 3D lidars, and enabling faster reaction times for collision avoidance through GPU acceleration. The NVIDIA Isaac ROS toolkit offers a low-latency visual 3D SLAM package~\cite{cuvslam} and 
uses GPU acceleration 
with nvblox~\cite{nvblox}, thus dramatically improving real-time robot navigation.
Together, these models have fundamentally transformed the capabilities of perception, planning, navigation, and manipulation.

\section{Experimental Setup}
\label{sec:setup}

Our setup can be broken down into the robotic platforms (\xref{subsec:robots}), and the compute and network infrastructure (\xref{subsec:infra}).

    


\begin{figure}[t]
    \centering

    \begin{subfigure}[t]{0.44\columnwidth}
        \centering
        \includegraphics[width=\linewidth]{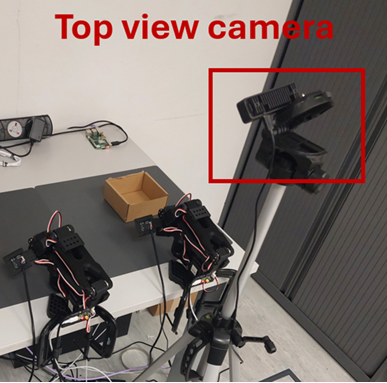}
        \caption{\small Bimanual SO-101 arm}
        \label{fig:so101}
    \end{subfigure}
    \hfill
    \begin{subfigure}[t]{0.4\columnwidth}
        \centering
        \includegraphics[width=\linewidth]{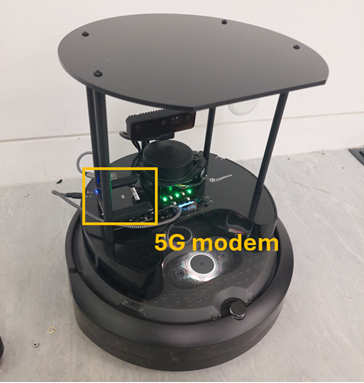}
        \caption{\small TurtleBot 4}
        \label{fig:turtlebot}
    \end{subfigure}

    \vspace{0.5em}

    \begin{subfigure}[t]{0.45\columnwidth}
        \centering
        \includegraphics[width=\linewidth]{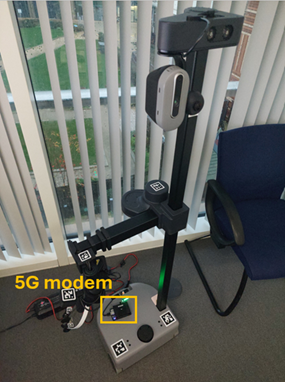}
        \caption{\small Stretch 3}
        \label{fig:stretch}
    \end{subfigure}
    \hfill
    \begin{subfigure}[t]{0.35\columnwidth}
        \centering
        \includegraphics[width=\linewidth]{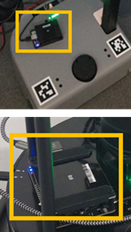}
        \caption{\small 5G modems}
        \label{fig:modems}
    \end{subfigure}

    \caption{\small Robotic platforms used for evaluating workloads.}
    \label{fig:robots}
    \vspace{-3mm}
\end{figure}


\subsection{Robotic platforms}
\label{subsec:robots}

Our goal in selecting robots for our experiments was to span key points in the design space, including robotic functionalities (e.g., manipulation, locomotion), hardware capabilities (e.g., sensors) and cost. We focused on robots with open software stacks and full programmability, allowing us to build on community-developed models. We selected three robotic platforms: Stretch 3, TurtleBot 4, and SO-101 (Figure \ref{fig:robots}). Together, these platforms cover a range of costs, onboard compute, battery power, and functional abilities.

\noindent{\bf 1) Stretch 3:}
Stretch 3 is an open-source mobile manipulator (Figure~\ref{fig:stretch}), designed for operation in homes and workplaces. It is equipped with a 7-DoF gripper, two RGB-D cameras (Intel RealSense D435i mounted on the head and Intel RealSense D405 on the gripper), a differential-drive base, and a 2D LiDAR (SLAMTEC RPLIDAR A1). It comes with an Intel NUC 12 compute and two 12V batteries providing $\sim$216 Wh combined. Stretch provides an open-source software stack, Stretch AI~\cite{stretchai}, with models for perception, planning, navigation, and manipulation. 
The cost of the robot is \$25K.

\noindent{\bf 2) TurtleBot 4:}
TurtleBot 4 is an open-source mobile robotics platform built on the iRobot Create 3 base (Figure~\ref{fig:turtlebot}). It includes a Luxonis OAK-D Pro RGB-D camera and a SLAMTEC RPLIDAR A1, both connected to a Raspberry Pi 4. In contrast to Stretch 3, it does not include a gripper, and is therefore suited for search and surveillance tasks but not manipulation. The platform includes a 26 Wh battery (14.4 V nominal). With widespread community support, it supports many navigation packages \cite{turtlebot4}, and also allows open-source models and algorithms to be easily adapted. Its cost is \$2K.

\noindent{\bf 3) Bimanual SO-101:}
The SO-101 is a robotic arm developed by RobotStudio and Hugging Face. It is suited for manipulation tasks, such as pick-and-place, and can be mounted on a mobile base, such as the LeKiwi platform \cite{sigrobotics_lekiwi_2025}. Each arm offers 6 degrees of freedom. The arm includes a Raspberry Pi-based wrist camera and connects to external compute, such as Raspberry Pi 5, through USB-C. For our experiments, we used a static bimanual configuration consisting of two camera-equipped SO-101 arms (Figure~\ref{fig:so101}), along with an additional Luxonis OAK-D camera mounted on a tripod to provide a top-down view. The arms are integrated with the LeRobot SDK~\cite{lerobot} for data collection, training, and evaluation using state-of-the-art VLA models. Each arm costs $\sim$\$300.


\subsection{Compute \& network infrastructure}
\label{subsec:infra}

\begin{figure}[t]
    \centering
    \includegraphics[width=1.0\linewidth]{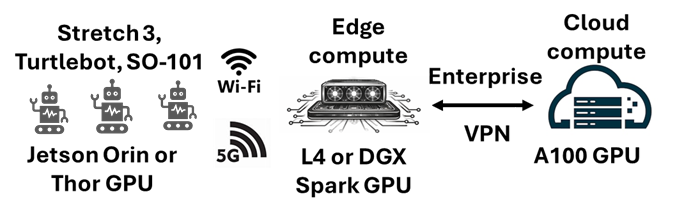}
    \caption{\small 
    Experimental setup of robots, edge compute with Wi-Fi and 5G connectivity, and cloud compute.
    \label{fig:xavier-arch}
    }
    \vspace{-3mm}
\end{figure}

The architecture of our testbed (Figure~\ref{fig:xavier-arch}), consists of robot, edge, and cloud components, as well as their connectivity.

\subsubsection{\bf Robot onboard compute:} Driven by recent trends of small form-factor and energy efficient options for robotics GPU acceleration, we leveraged (1) an Nvidia Jetson Orin AGX with 32GB of unified memory and (2) an Nvidia Jetson Thor with 128GB of unified memory as the primary onboard compute options.
We also experimented with (3) an Nvidia Jetson Orin Nano as a lighweight alternative, but, with only 8GB of unified memory, few of the models could fit.
We extended the robots to include Nvidia Jetsons as additional onboard compute.
To measure the power consumption of the onboard compute boards, we used a smart plug (Shelly Plug Gen 4), that allows us to export power measurements in real time over the network using a network API~\cite{shelly_plug_gen4_docs}.

\subsubsection{\bf Edge compute:}
To offload to the edge, we considered: (1) an HPE server with an Intel Xeon 6338N CPU, an Nvidia L4 GPU with 24GB of memory, and an Intel E810 NIC at 25 Gbps; (2) an Nvidia DGX Spark workstation with 128GB of {\em unified} memory and ConnectX-7 NIC at 200 Gbps.

The aforementioned options for compute onboard and at the edge capture different representative points in the design space of cost, power, and performance.

\subsubsection{\bf Cloud compute:} To offload to the cloud, we used an Azure VM with an AMD EPYC 7V13 CPU, an A100 GPU with 80GB of memory, and a Mellanox NIC at 100 Gbps.

\subsubsection{\bf Connectivity:}
The robots have wireless connectivity using both Wi-Fi and 5G. 
We deployed a Netgear Business WAX602 Wi-Fi 6 access point, and we equipped the robots with Wi-Fi 6 AX900 USB adapters.
For 5G, we deployed a private 5G network using srsRAN~\cite{srsran_project}, open5GS, and a Foxconn RPQN 7800 radio unit, and equipped the robots with 5G modems 
(Figure~\ref{fig:modems}).
The edge compute was connected to the cloud VM via a FortiGate firewall and VPN. 
\section{Workloads Under Study and Methodology}
\label{sec:methodology}

\begin{table*}[]
\footnotesize
\begin{tabular}{|c|c|c|c|c|p{0.35cm}|p{0.4cm}|p{0.45cm}|}
\cline{1-8}
\textbf{}                                                                                                 \textbf{Class}                                                      & \textbf{Workload}                                           & \textbf{Task description}                                                                                                                                                                                 & \textbf{Operations involved}                                                                                 & \textbf{Embodiment} & \textbf{BW} & \textbf{Lat} & \textbf{Cmp} \\ \hline
\rowcolor[HTML]{C0C0C0} 
\multicolumn{1}{|c|}{\multirow{2}{*}{\cellcolor[HTML]{C0C0C0}\shortstack{Semantic understanding \\ \& long-term planning}}} & VLMaps~\cite{vlmaps}          & \begin{tabular}[c]{@{}c@{}}The robot updates a semantic map \\ with new information as it moves \\ from point A to point B of the lab.\end{tabular}                                                                   & \begin{tabular}[c]{@{}c@{}}Perception \& navigation  \end{tabular}             & TurtleBot           & H & M & H \\
\cline{2-8} 
\rowcolor[HTML]{C0C0C0} 
\multicolumn{1}{|c|}{\cellcolor[HTML]{C0C0C0}}                                                                                                                  & GraphEQA~\cite{grapheqa}                                         & \begin{tabular}[c]{@{}c@{}}The robot is asked to check if the \\ door of the lab has been left open.\end{tabular}                                                                            & \begin{tabular}[c]{@{}c@{}}Perception, planning, \\ \& navigation \end{tabular}   & Stretch 3           & L & M & H \\ 
 \hline
\rowcolor[HTML]{EFEFEF} 
\multicolumn{1}{|c|}{\cellcolor[HTML]{EFEFEF}\begin{tabular}[c]{@{}c@{}}Multi-step mobile \\ manipulation\end{tabular}}                                                                                                                  & $\pi_{0.5}$~\cite{pi0.5}                                                & \begin{tabular}[c]{@{}c@{}}The robot is tasked with handing \\ over a tape from one arm to the \\ other and to place it in a box \\ within a time window of 70s.\end{tabular}                                                                 & \begin{tabular}[c]{@{}c@{}}Perception, planning, \\ \& manipulation \end{tabular} & SO-101              & H & L & H \\
\cline{2-8} 
\rowcolor[HTML]{EFEFEF} 
\multicolumn{1}{|c|}{\cellcolor[HTML]{EFEFEF}}                             & DreamZero~\cite{ye2026world}                                                  & \begin{tabular}[c]{@{}c@{}}The robot is tasked with grabbing \\ a frying pan from the hob.\end{tabular}                                                                  & \begin{tabular}[c]{@{}c@{}}Perception, planning, \\ \& manipulation\end{tabular} & \begin{tabular}[c]{@{}c@{}} Simulation \\ (video rec.) \end{tabular}   & H & L & H 
\\ 
\hline
\rowcolor[HTML]{9B9B9B} 
\multicolumn{1}{|c|}{\cellcolor[HTML]{9B9B9B}Collision-free navigation}                                                                                                   & \begin{tabular}[c]{@{}c@{}}RTAB-Map~\cite{rtabmap} \\ \&  nvblox~\cite{nvblox}\end{tabular} & \begin{tabular}[c]{@{}c@{}}The robot is tasked to navigate \\ to some location, which requires it \\ to turn around a corner and avoid \\ a previously unmapped obstacle.\end{tabular} & \begin{tabular}[c]{@{}c@{}}Perception \& navigation\end{tabular}                                                                                  & TurtleBot           & H & L & M \\ \hline
\end{tabular}
\caption{\small Robotic workloads considered for the experimental evaluation characterized by bandwidth (\textbf{BW}), latency (\textbf{Lat}), and compute (\textbf{Cmp}) requirements as low (L), medium (M), or high (H).}
\label{tab:eval-robotic-workloads}
\end{table*}

We construct a suite of evaluation benchmarks consisting of key building blocks of mobile robotic manipulation (Table~\ref{tab:eval-robotic-workloads}). 
\begin{enumerate*}[label=(\roman*)]
    \item semantic understanding of the environment and long-term planning (\xref{subsec:workload-semantic}), 
    \item safe (i.e., collision-free) navigation  (\xref{subsec:workload-navigation}), and
    \item multi-step manipulation involving sub-tasks (\xref{subsec:workload-manipulation}).
\end{enumerate*} 
For the canonical task of “finding all the dirty dishes in the office and bringing them to the kitchen sink”, the robot needs to understand the environment it is in and make a plan (operation (i)), safely navigate to key locations (operation (ii)), and then manipulate the objects (operation (iii)).


\subsection{Semantic mapping and planning}
\label{subsec:workload-semantic}

Semantic mapping provides understanding of the robot's environment, which in turn is used by a planner to devise the steps to execute mobile robotic manipulation tasks. 

\subsubsection{\bf VLMaps:}
As described in \xref{subsec:foundation}, foundation models have enabled rich 3D scene modeling. VLMaps \cite{vlmaps} is one such popular approach that allows for continuously updating an open-vocabulary semantic map. Like most modern semantic mappers, it builds a 3D point cloud, where each point is a vector in a joint image-text embedding space. We improved the VLMaps implementation to use the SAM segmentation model~\cite{sam}, CLIP~\cite{clip} to generate image embeddings, and Depth Anything \cite{depthanything} to improve depth estimates. These modifications together substantially improved accuracy of the 3D map when we deployed it on the TurtleBot robot (\xref{subsec:robots}). The map generated by VLMaps is semantically rich and is used in conjunction with a language model for open-vocabulary navigation tasks. VLMaps requires high bandwidth image streams and high compute to generate its map. Map updates are expected with medium latency (seconds to minutes).

\subsubsection{\bf GraphEQA:}
Building upon 3D mapping, GraphEQA \cite{grapheqa} combines mapping {\em as well as} planning. While similar in spirit to VLMaps, GraphEQA's mapping has a key differentiator: it uses a 3D semantic scene graph, in which nodes correspond to spatial elements such as rooms and objects, and edges capture their geometric and semantic relationships. The scene graph allows GraphEQA to answer questions, such as ``is the microwave in the kitchen left open?''. To generate its answer, GraphEQA uses a multimodal LLM planner (such as Qwen-VL-2.5-3B~\cite{qwen2.5}) that interprets natural language queries and plans the steps needed to answer the above question, such as navigating to the kitchen, identifying the microwave, and understanding the state of its door. 
GraphEQA is supported by the Stretch AI framework of the Stretch robot (\xref{subsec:robots}). 
It requires low bandwidth because it only processes videos occasionally, but requires high compute when a question is asked. It tolerates medium response latency (seconds).



\subsection{Collision-free navigation}
\label{subsec:workload-navigation}

Navigation in mobile robotic manipulation applications allows robots to move to the locations of objects of interest.

\subsubsection{\bf RTAB-Map \& nvblox:}
\begin{figure}[t]
    \centering
    \includegraphics[width=0.7\linewidth]{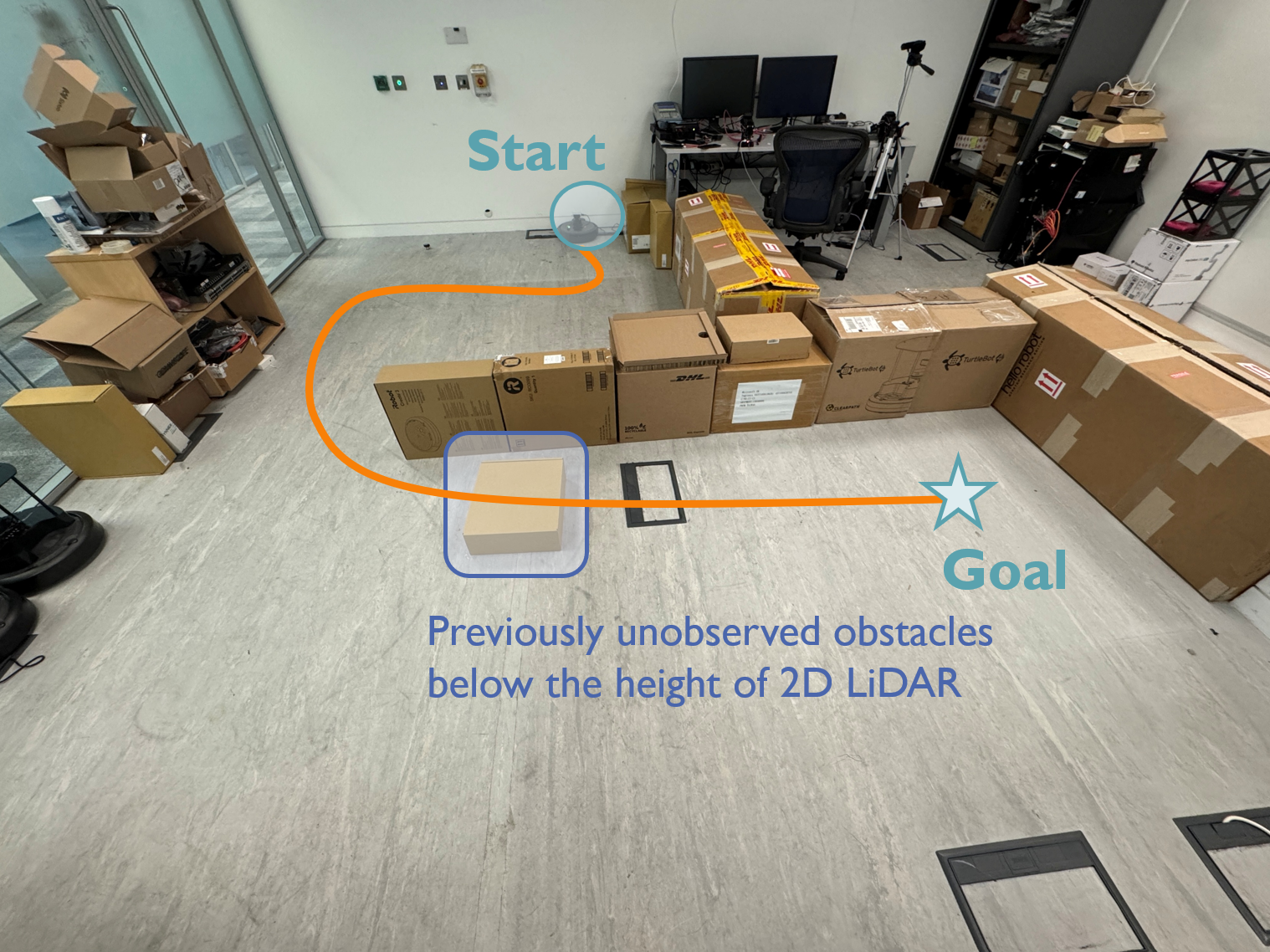}
    \caption{\small
    Trajectory of TurtleBot for navigation task and unmapped hidden obstacle.
    \label{fig:nvblox_setup}
    }
    \vspace{-3mm}
\end{figure}

We evaluate SOTA SLAM methods, which consist of two parts: localization and mapping (\xref{subsec:overview}). For localization, we employ RTAB-Map \cite{rtabmap}--a widely used SLAM method that fuses RGB-D, Stereo, and LiDAR sensor inputs. 
Although RTAB-Map can also generate a map of the environment, map updates can take many seconds, which is not sufficient in highly dynamic environments or even some static environments, depending on the operating speed of the robot. In our task, the TurtleBot has a partially known map and is tasked with turning around a corner with an unmapped obstacle (Figure~\ref{fig:nvblox_setup}), which is not visible to the robot until after it has turned, so it must quickly detect the obstacle and update the map. RTAB-Map does not update its map quickly enough to avoid collision, so we use a GPU-based occupancy mapping method, nvblox \cite{nvblox}, which can update the map an order of magnitude faster.

\subsection{Mobile manipulation}
\label{subsec:workload-manipulation}

Finally, manipulation models allow robots to interact with objects by picking and placing them.

\subsubsection{\bf $\mathbf{\pi_{0.5}}$:}
Foundation models have been used to design VLAs to enable manipulation skills generalizing across tasks and robot embodiments (\xref{subsec:foundation}). The most widely used of these models are $\pi_0$ \cite{pi0} and its successor, $\pi_{0.5}$ \cite{pi0.5}. These VLAs combine a VLM backbone, PaliGemma \cite{paligemma}, with a smaller action expert diffusion model that processes robot states and actions. This architecture allows a single model to generalize across robots with differing action spaces to perform a wide range of tasks. Using the LeRobot framework \cite{lerobot}, we fine-tune the $\pi_{0.5}$ model on the bimanual SO-101 robotic arm (\xref{subsec:robots}; Figure \ref{fig:robots}c) to {\em hand over a tape from one arm to the other, and then place it in a box}. Such coordinated manipulation is representative of real-world scenarios in manufacturing and enterprises, for moving tools, hazardous objects, or cups and cans. We deploy the fine-tuned $\pi_{0.5}$ model 
using action chunks of 50 actions and an execution horizon of 20 actions. We also use the state-of-the-art real-time chunking (RTC) algorithm for tolerating high inference and network latencies~\cite{rtc}, but we still expect an end-to-end inference latency of 100--200 ms.

\subsubsection{\bf DreamZero:} 
Another emerging class of robot foundation models, called World Action Models (WAMs), tightly couple action generation with world modeling. DreamZero~\cite{ye2026world} is a representative WAM that builds on a large pretrained video diffusion backbone to jointly predict future visual observations and robot actions conditioned on language and robot state. By aligning actions with predicted video futures, DreamZero leverages spatiotemporal priors from web-scale videos, enabling generalization to unseen tasks, spaces, and robots without explicit planning. Due the lack of support of DreamZero for the onboard GPUs that we studied, we use it in simulation, using video feeds of a robot tasked with grabbing a frying pan from the hob. 
Our evaluation of this workload focused on its compute and performance requirements.



\section{Analysis of Onboard Compute Options}
\label{sec:on_board}

We leverage our setup (\xref{sec:setup}) and workloads (\xref{sec:methodology}) to investigate the resource demands of robotic applications. 
We run each workload 
on three onboard GPUs: Jetson Orin AGX, Thor, and Nano. We report {\em i)} memory consumption (\xref{subsec:onboard-memory}), {\em ii)} end-to-end task execution time and/or accuracy (\xref{subsec:onboard-perf}), and {\em iii)} power consumption (\xref{subsec:onboard-power}). We find that the demanding requirements of robotics workloads make onboard GPUs inadequate.





\subsection{Memory usage}
\label{subsec:onboard-memory}

Given the unified memory architecture of platforms like the Orin and the Thor, the CPU and GPU tasks have to contend for the same shared memory. We therefore measure both the CPU and GPU memory for all workloads using the monitoring tools provided by Nvidia (\verb|jetson-stats| ~\cite{jetson_stats} for the Jetson Orin AGX, and \verb|nvidia-smi|~\cite{nvidia_smi} for the Jetson Thor), as well as the Nsight Systems  tool~\cite{nvidia_nsight_systems}, for GPU and CPU behavior.

\begin{figure}[t]
    \centering
    \includegraphics[width=\columnwidth]{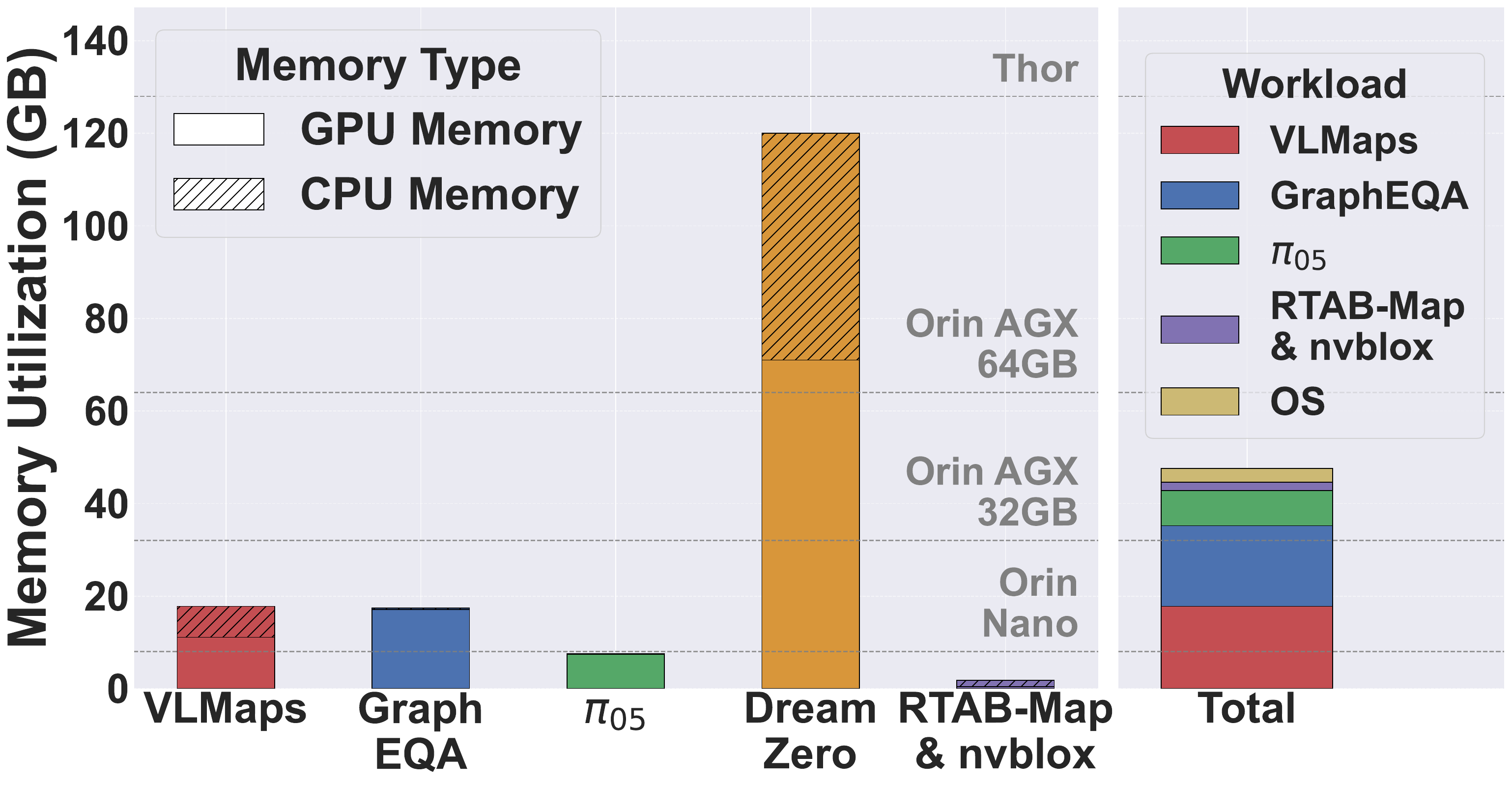}
    \caption{\small GPU and CPU memory consumption of workloads.}
    \label{fig:memory-consumption}
\end{figure}

As depicted in Figure~\ref{fig:memory-consumption}, we observe that all the workloads under study (\xref{sec:methodology}) require several gigabytes of memory, ranging from approximately 2GB for the RTAB-Map \& nvblox workload, to more than 120GB for DreamZero.



As we can see in Figure~\ref{fig:memory-consumption}, even without taking DreamZero into account, the full stack of workloads, including semantic mapping and planning, mobile manipulation, and collision-free navigation, plus the OS, requires a minimum of approximately 50GB of memory. 
This makes it challenging to simultaneously deploy several workloads for the robot on memory-constrained platforms, like the Nano or the 32GB Orin. For reference, Figure~\ref{fig:memory-consumption} includes the memory capacities of the GPUs typical for onboard robotics, showing that the demands are stringent {\em even for one workload}. While one could swap models into memory on demand, the overheads of many seconds will be prohibitive for real-time mobile manipulation.

\subsection{Task performance}
\label{subsec:onboard-perf}

We next evaluate the 
accuracy and execution time of the workloads of Table~\ref{tab:eval-robotic-workloads} on the onboard GPU options of Jetson Thor, Orin AGX, and Nano (where applicable).
For {\em comparative reference}, we use the accuracy and execution time achieved when offloading the workloads to the A100 GPU VM, with an ideal maximum round-trip network latency of less than 15ms. 
We repeat each task described in Table~\ref{tab:eval-robotic-workloads} twenty times to report our results (Figures~\cref{fig:vlmaps-results,fig:grapheqa-results,fig:nvblox-results,fig:pi-results}).

\subsubsection{\bf VLMaps:} 
\begin{figure}[t]
  \centering
  \begin{subfigure}[t]{0.4\textwidth}
    \centering
    \includegraphics[width=0.8\linewidth]{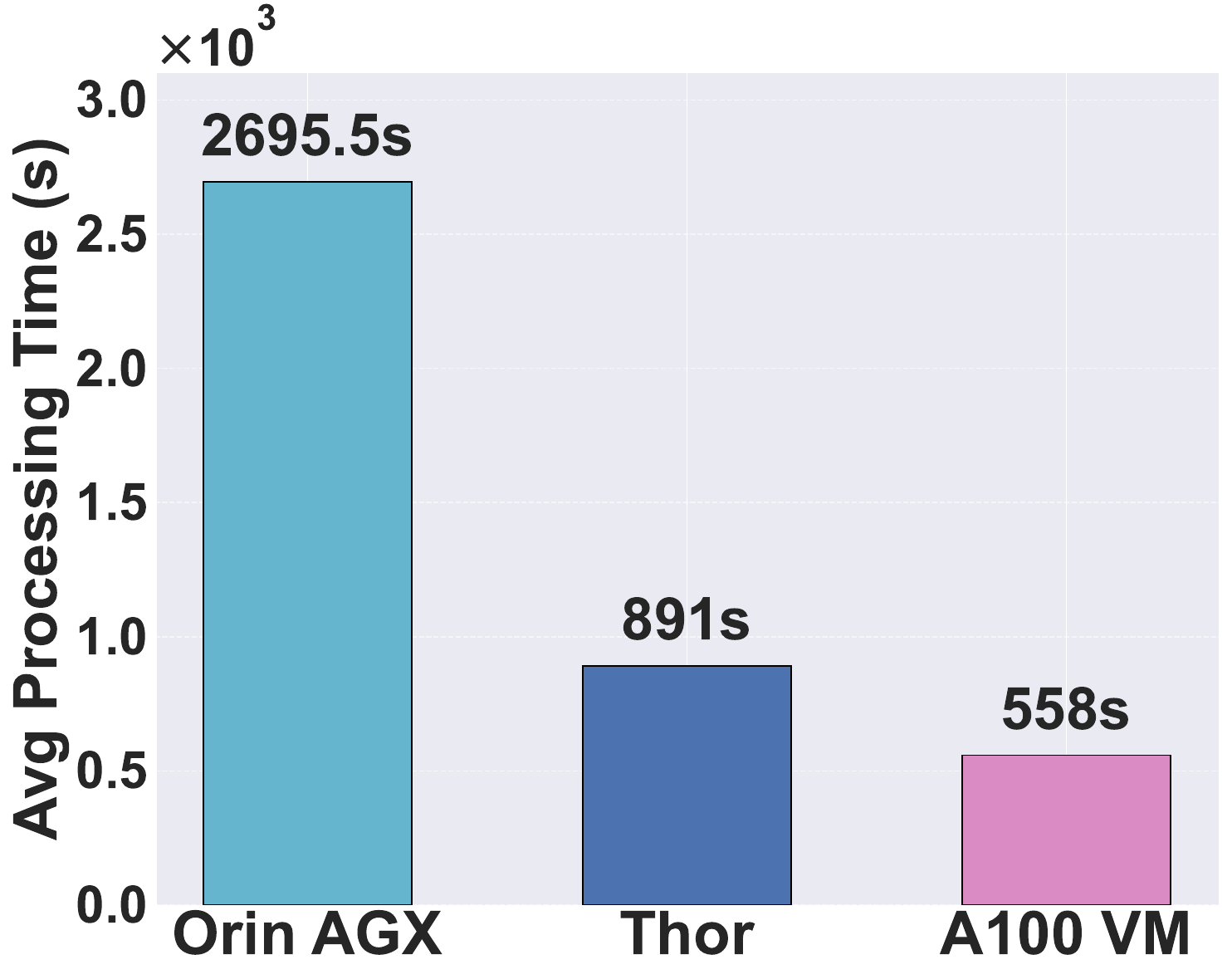}
    \caption{\small Average processing time for 450 frames.}
    \label{fig:vlmaps_processing_time}
  \end{subfigure}


  \begin{subfigure}[t]{0.4\textwidth}
    \centering
    \includegraphics[width=0.8\linewidth]{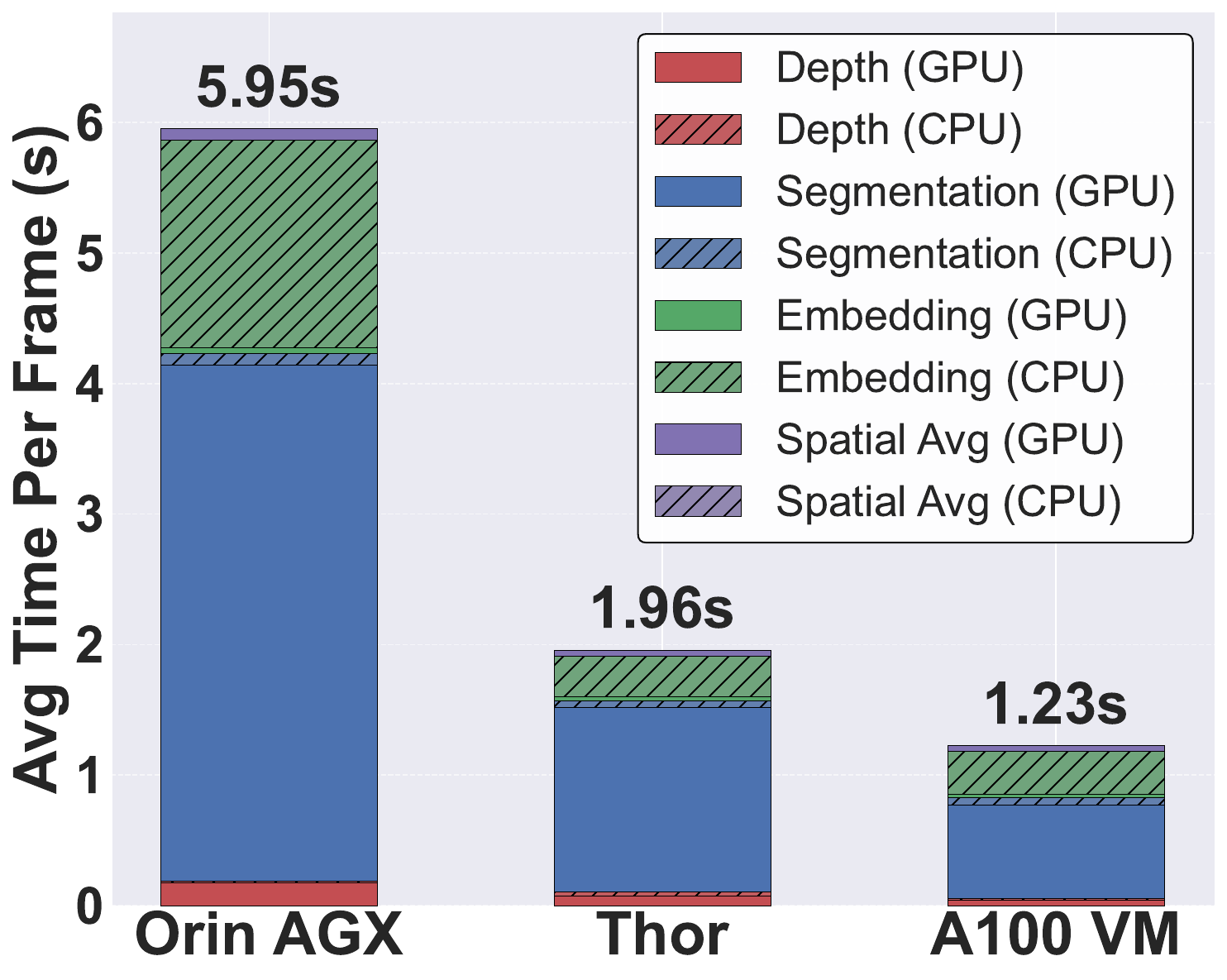}
    \caption{\small CPU and GPU time for processing a single frame.}
    \label{fig:vlmaps_breakdown}
  \end{subfigure}

  \caption{\small VLMaps execution time across compute platforms.}
  \label{fig:vlmaps-results}
\end{figure}

Recall from \xref{sec:methodology} that VLMaps is used to generate a rich 3D map of a robot's environment. We measure the time for map generation with 450 frames (30s of video at 15 FPS). Note that the lower the generation time, the more current the robot's understanding of its environment, which is vital in many dynamic situations. 
As Figure~\ref{fig:vlmaps_processing_time} shows, compared to our reference execution time on an A100, the execution time jumps by 59\% with the Thor and 383\% with the Jetson Orin AGX--a slowdown that can degrade the utility of the semantic map and be detrimental to the robot's performance in frequently changing environments. 
Upon deeper inspection of the time and task breakdown bars in Figure~\ref{fig:vlmaps_breakdown}, the frame processing is primarily a GPU-bound operation, with the majority of the time spent on the GPU task of segmentation using the SAM model (see \xref{subsec:workload-semantic}).
Interestingly, the second biggest contributor to the total frame processing time is the CPU part of the embedding stage, which involves large data copying.
Nonetheless, this CPU task performs similarly on the Thor and the A100 VM, making the GPU task of segmentation the main contribution to the spike in the overall processing time.


\subsubsection{\bf GraphEQA:} 
\begin{figure}[t]
  \centering
  \begin{subfigure}[t]{0.4\textwidth}
    \centering
    \includegraphics[width=0.8\linewidth]{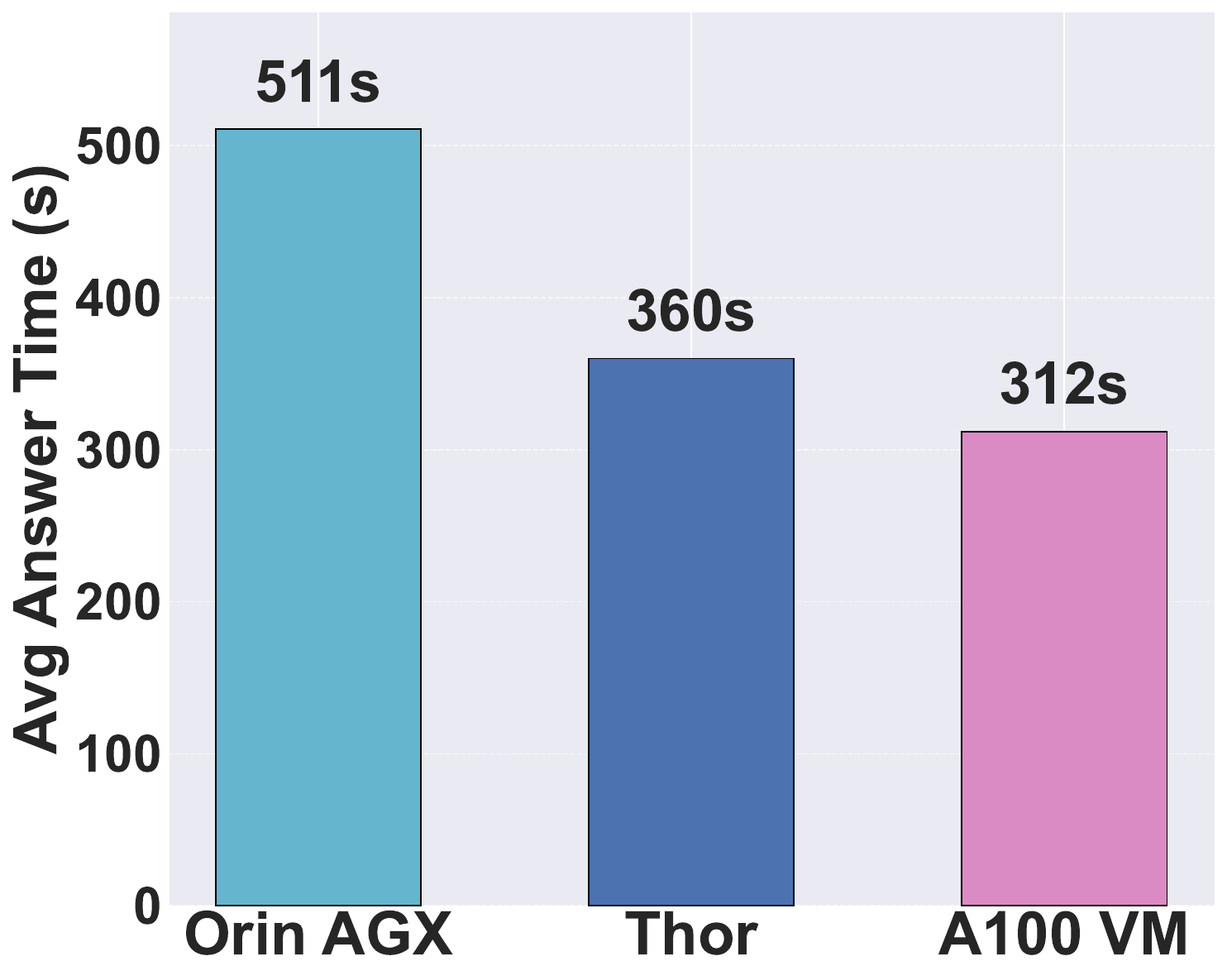}
    \caption{\small Average answer time for GraphEQA.}
    \label{fig:graph_eqa_time}
  \end{subfigure}

  \vspace{0.5em}

  \begin{subfigure}[t]{0.45\textwidth}
    \centering
    \includegraphics[width=0.8\linewidth]{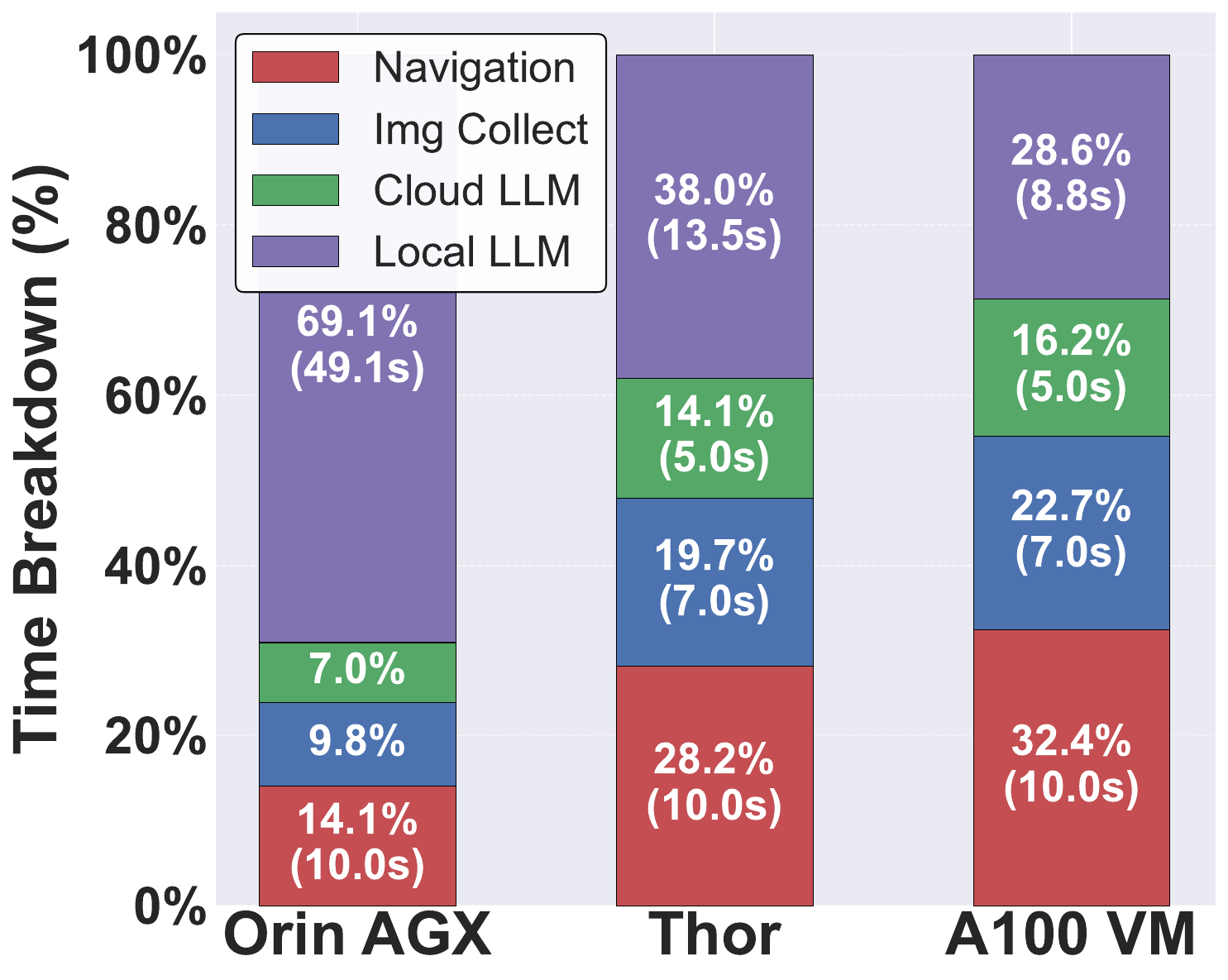}
    \caption{\small Breakdown of total task time per GraphEQA task.}
    \label{fig:grapheqa_breakdown}
  \end{subfigure}

  \caption{\small GraphEQA execution time across compute platforms.}
  \label{fig:grapheqa-results}
  \vspace{-3mm}
\end{figure}

Recall from \xref{subsec:workload-semantic} that GraphEQA uses its 3D map to construct and execute a workflow of steps for the robot, in order to answer questions about the environment.   
As seen in Figure~\ref{fig:graph_eqa_time}, the end-to-end execution time of the task varies significantly across GPU platforms, with the A100 VM requiring on average 312s, the Thor 360s, and the Orin AGX 511s. 
For spatial exploration, interactivity and small user waiting times are important for the utility of the application, and the inflation in Figure~\ref{fig:graph_eqa_time} can be sizable. 
As shown in the breakdown in Figure~\ref{fig:grapheqa_breakdown}, the main contributor to this time increase is the GPU inference time for the local VLM, which is used to identify featured object names from the collected image observations. With VLM model sizes predicted to increase, the GPU compute will be the expected bottleneck.


\subsubsection{\bf RTAB-Map \& nvblox:}
\begin{figure}[t]
  \begin{subfigure}[t]{0.22\textwidth}
    \centering
    \includegraphics[width=\linewidth]{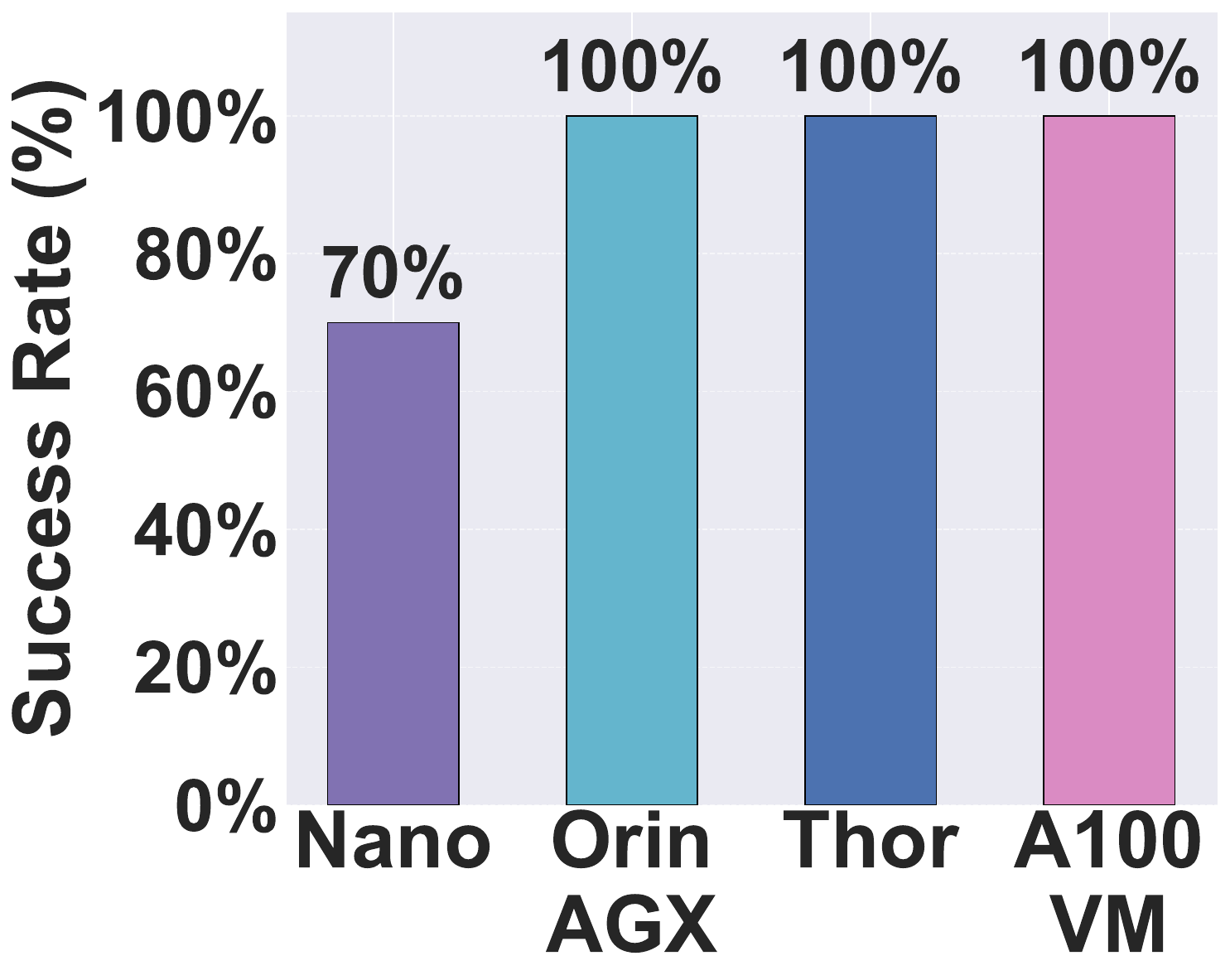}
    \caption{\small Accuracy of obstacle avoidance with nvblox.}
    \label{fig:obstacle_avoidance_accuracy}
  \end{subfigure}\hspace{.1in}
  \begin{subfigure}[t]{0.22\textwidth}
    \centering
    \includegraphics[width=\linewidth]{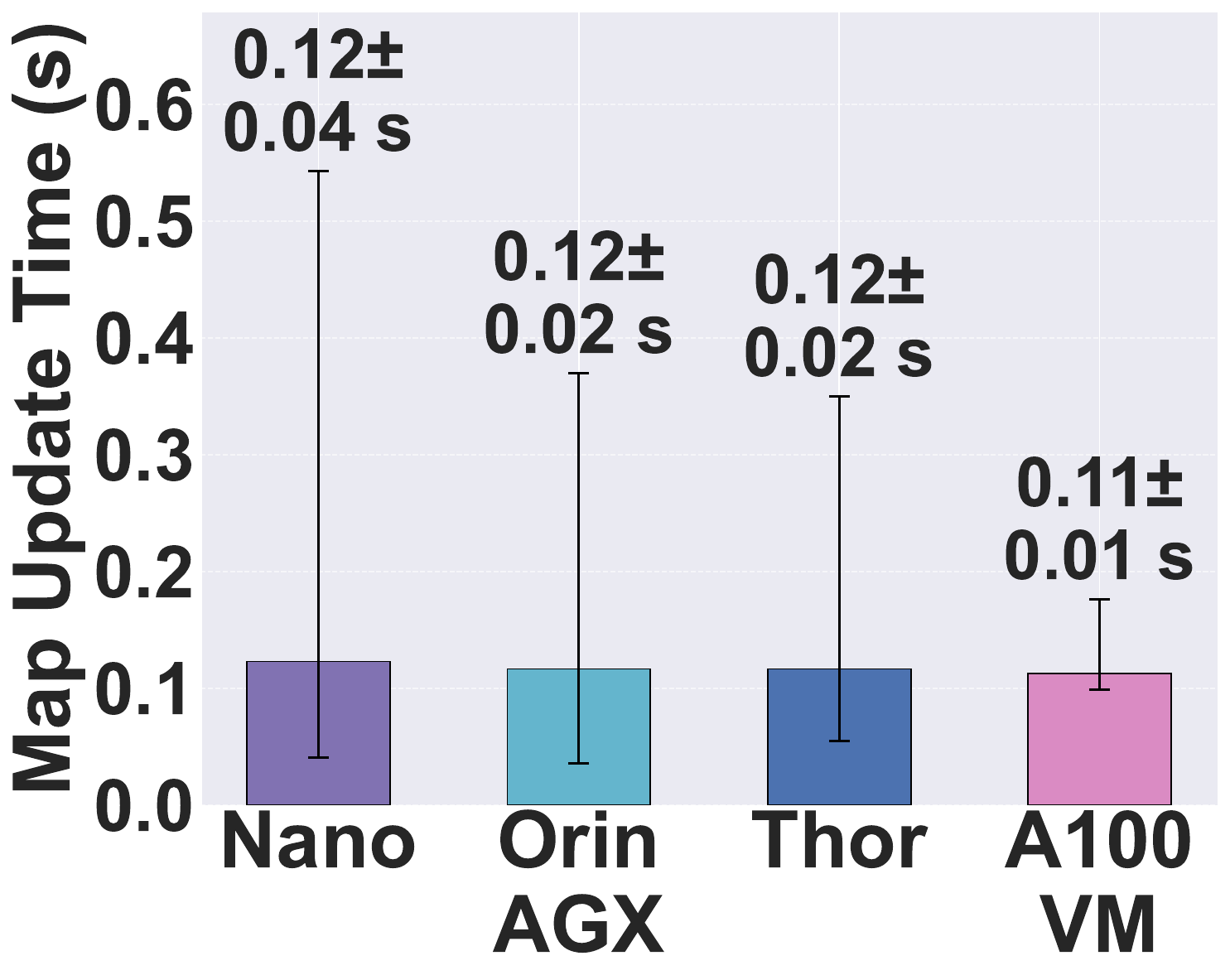}
    \caption{\small Nvblox map update duration for obstacle detection.}
    \label{fig:map_update_rate}
  \end{subfigure}
  \caption{\small Nvblox success rates across compute platforms.}
  \label{fig:nvblox-results}
  \vspace{-3mm}
\end{figure}

Given the small memory footprint of the nvblox component, we evaluate its performance on the more lightweight Jetson Nano, in addition to the Thor and Orin AGX.
As observed in Figure~\ref{fig:obstacle_avoidance_accuracy}, the task success rate is 100\% on all platforms except the Nano, for which it drops to 70\%.
This indicates that, despite the lightweight nature of the navigation workload, higher-end onboard compute might still be required to ensure high reliability, considering that collision-free navigation is a safety-critical operation.
Inspection of the failed runs reveals that the accuracy reduction is due to delays in detecting and updating unmapped occupancies (\xref{subsec:workload-navigation}), shown in Figure~\ref{fig:map_update_rate}.
While the average latency is approximately the same, the Nano sees significantly higher standard deviation, and even higher max latencies that can reach up to half a second.
One could mitigate this by slowing the robot, but in many scenarios (e.g., factory floors, warehouses), reduced speed directly translates to reduced productivity.

\begin{figure*}[t]
  \begin{subfigure}[t]{0.3\textwidth}
    \centering
    \includegraphics[width=\linewidth]{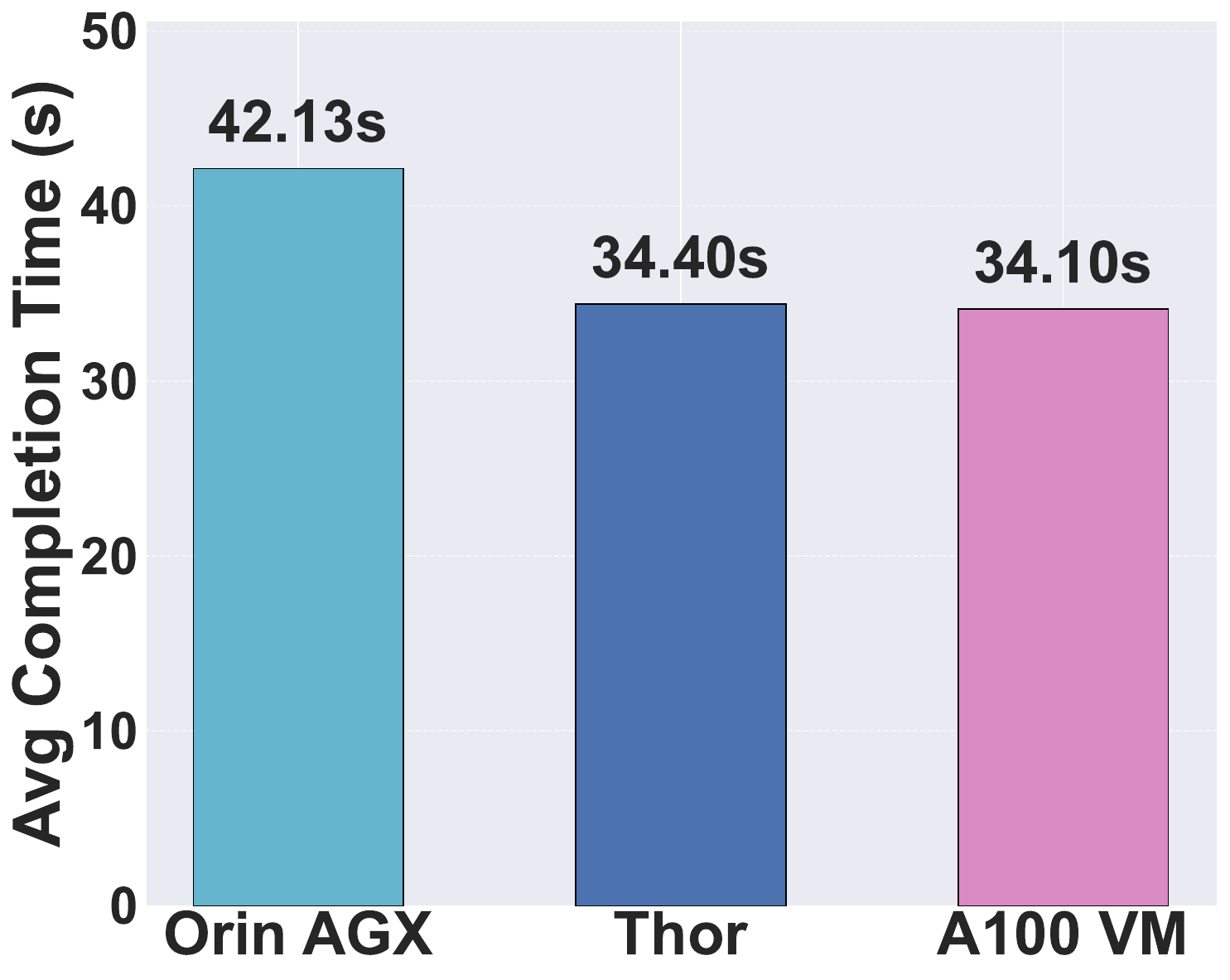}
    \caption{$\pi_{0.5}$ handover task execution time averaged across successful attempts.}
    \label{fig:vla_handover_time}
  \end{subfigure}\hspace{.1in}
  \begin{subfigure}[t]{0.3\textwidth}
    \centering
    \includegraphics[width=\linewidth]{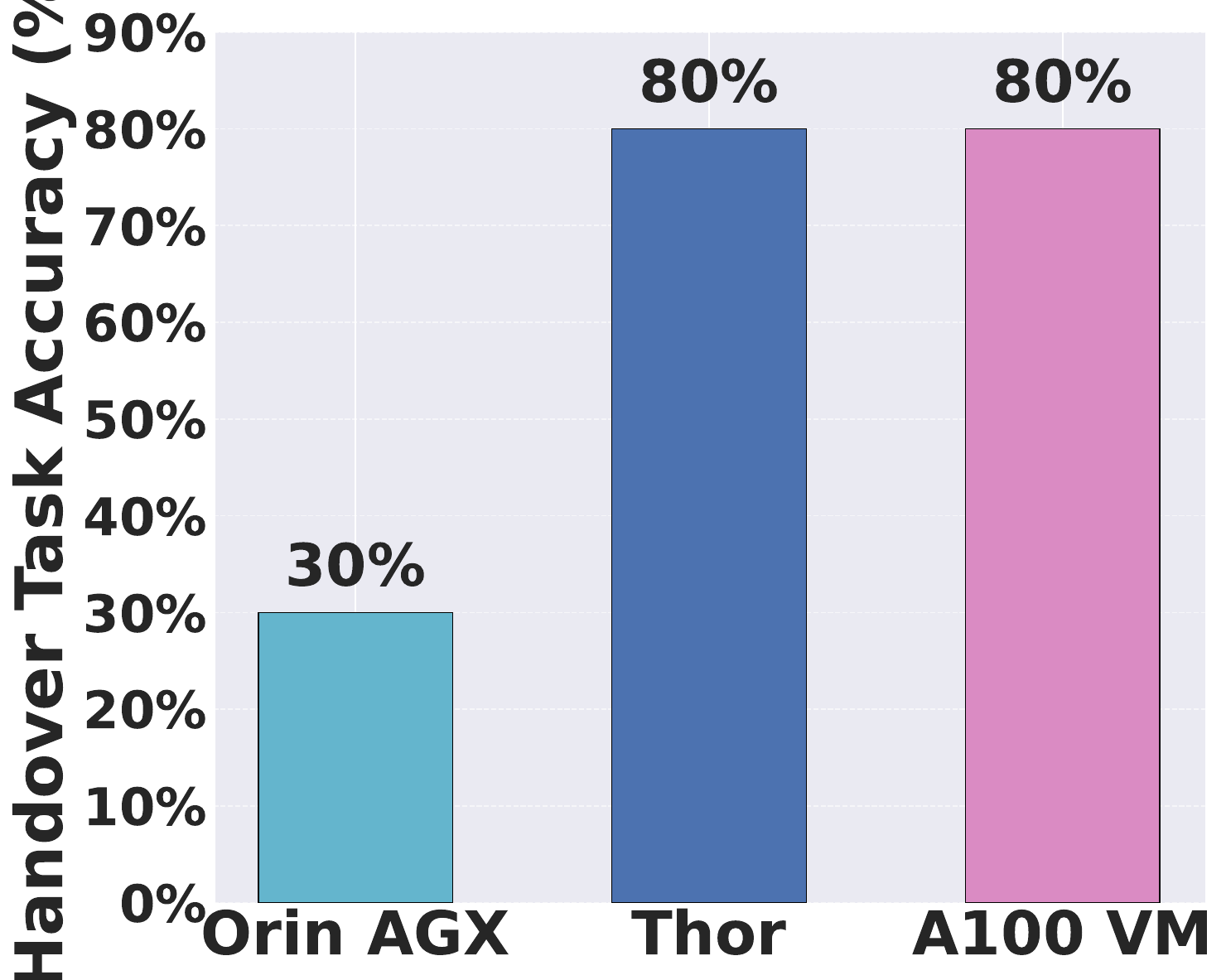}
    \caption{$\pi_{0.5}$ handover task accuracy.}
    \label{fig:vla_handover_accuracy}
  \end{subfigure}\hspace{.1in}
  \begin{subfigure}[t]{0.3\textwidth}
    \centering
    \includegraphics[width=\linewidth]{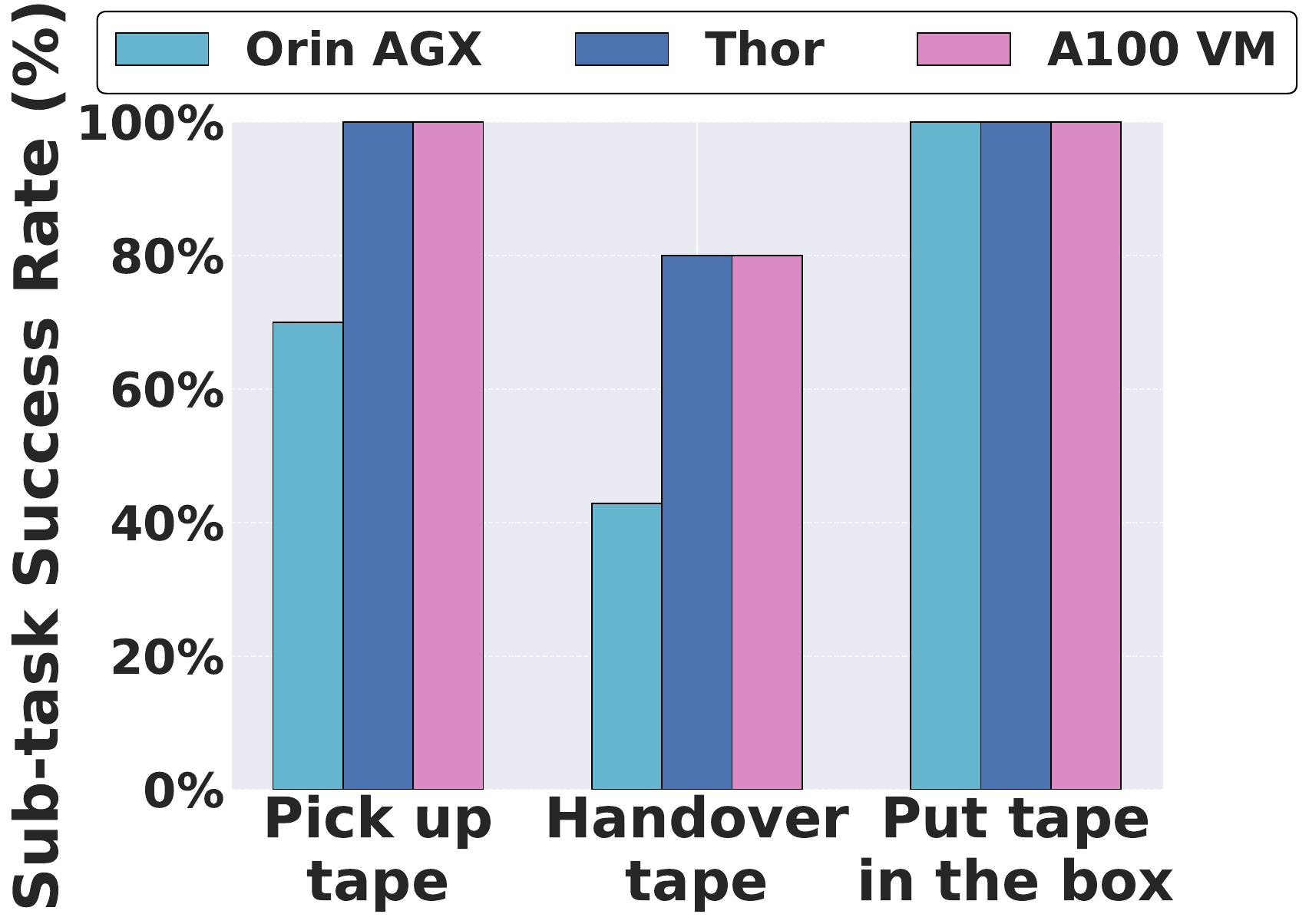}
    \caption{Success rate of sub-tasks conditioned on the success of the previous.}
    \label{fig:sub-task-vla-success-rate}
  \end{subfigure}
    \caption{\small $\pi_{0.5}$ execution times and success rates across compute platforms.}
  \label{fig:pi-results}
\end{figure*}

\subsubsection{\bf $\pi_{0.5}$:} For the handover task (\xref{subsec:workload-manipulation}), Figure~\ref{fig:vla_handover_time} shows the average end-to-end task execution time of successful attempts.
We observe that it is the same on the A100 and the Thor ($\sim$34s), and increases by $\sim$23\% on the Orin AGX ($\sim$42s). 
Crucially, the task accuracy (Figure~\ref{fig:vla_handover_accuracy}) on the A100 VM and the Thor is 80\%, while it drops to 30\% on the Orin.
This drop is a direct consequence of the latency for generation of action chunks: approximately 60ms for the A100, 144ms for the Thor and 440ms for the Orin AGX (see Figure~\ref{fig:inference_times_vla_wam} in Section~\ref{subsec:latency}). 
High latencies lead to the depletion of the action queue before a new action chunk is generated, leading to stop-and-go behavior and movement jerkiness that affects the accuracy of the task
~\cite{black2025real, tang2025vlash}.
We visually observed such jerkiness for the Orin AGX.
Also, we note that the latency affected the accuracy of the sub-tasks (i.e., pick up tape, handover, put in the box) differently. 
As shown in Figure~\ref{fig:sub-task-vla-success-rate}, failures manifested in the sub-tasks of picking up the tape and handing it over to the other arm. 
From our visual observations, this is due to the higher precision required by those sub-tasks, where mistakes of even a few centimeters can lead to dropping or failing to grab the tape.

\subsubsection{\bf DreamZero:} 
We evaluated DreamZero on simulations and focus on its execution time. 
We only consider the Thor and the A100, since the model does not fit on the Orin AGX.
The average inference time of DreamZero is 6.22s on the A100 and 19.27s on the Thor (see Figure~\ref{fig:inference_times_vla_wam} in ~\xref{subsec:latency}). 
Such latencies would also reduce task accuracy, as DreamZero requires sub-second inference, currently achievable only with high-end multi-GPU setups (2$\times$GB200)~\cite{ye2026world}, further underscoring the trend toward compute that exceeds onboard capabilities.
\newline

\noindent\fbox{\parbox{0.97\linewidth}{
\noindent\textbf{Takeaway \#1:} \textit{Modern robotics workloads require GPUs beyond what is typically fitted on board robots today, leading to expensive on-robot upgrades or offloading.}
}}

\subsection{Power consumption}
\label{subsec:onboard-power}

\begin{figure}[t]
    \centering
     \includegraphics[width=0.9\columnwidth]{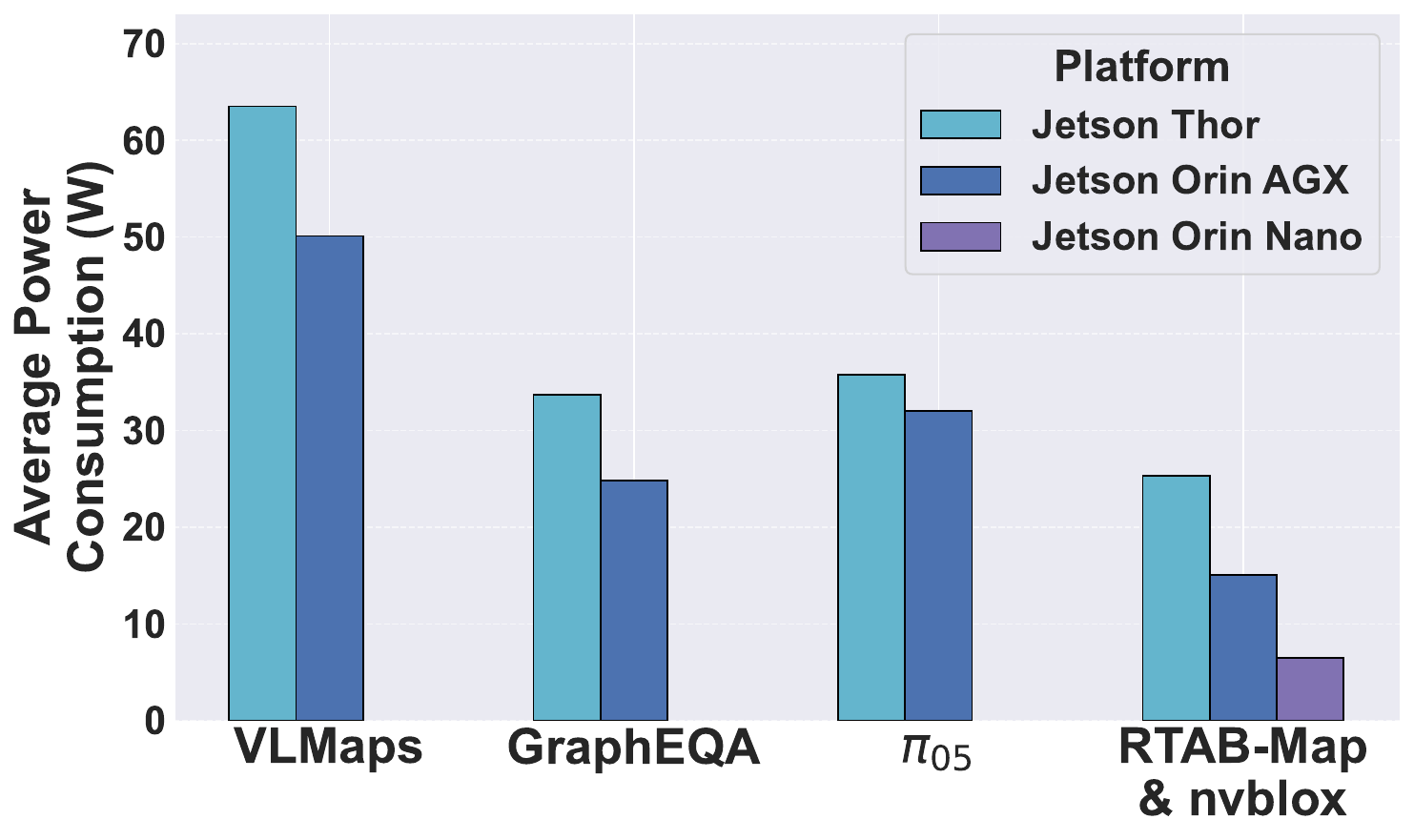}
    \caption{\small
    Power consumption of onboard compute when running the workloads of Table~\ref{tab:eval-robotic-workloads}.
    }
    \label{fig:avg_power_consumption}
    \vspace{-3mm}
\end{figure}

Another important dimension is the onboard compute power consumption and its effect on the battery lifetime of the robot. 
To this end, we measured the power consumption of our onboard compute platforms (Thor, Orin AGX, Nano), while running the various workloads of Table~\ref{tab:eval-robotic-workloads}, excluding DreamZero, which can only be hosted in server-grade GPUs. 
The results are summarized in Figure~\ref{fig:avg_power_consumption} and capture the average power consumption during the execution of each workload individually, including both busy and idle periods.
In the case of the Nano, we only report the power consumption for the RTAB-Map \& nvblox workload, as it is the only workload that can run on this platform.
From Figure~\ref{fig:avg_power_consumption} we observe that the VLMaps workload is the most demanding, followed by $\pi_{0.5}$, GraphEQA, and RTAB-Map \& nvblox. It is noteworthy that the Thor consumes only marginally more power compared to the Orin AGX, while exhibiting significantly better compute capabilities (\xref{subsec:onboard-perf}).


\begin{figure}[t]
    \centering
    \includegraphics[width=0.9\columnwidth]{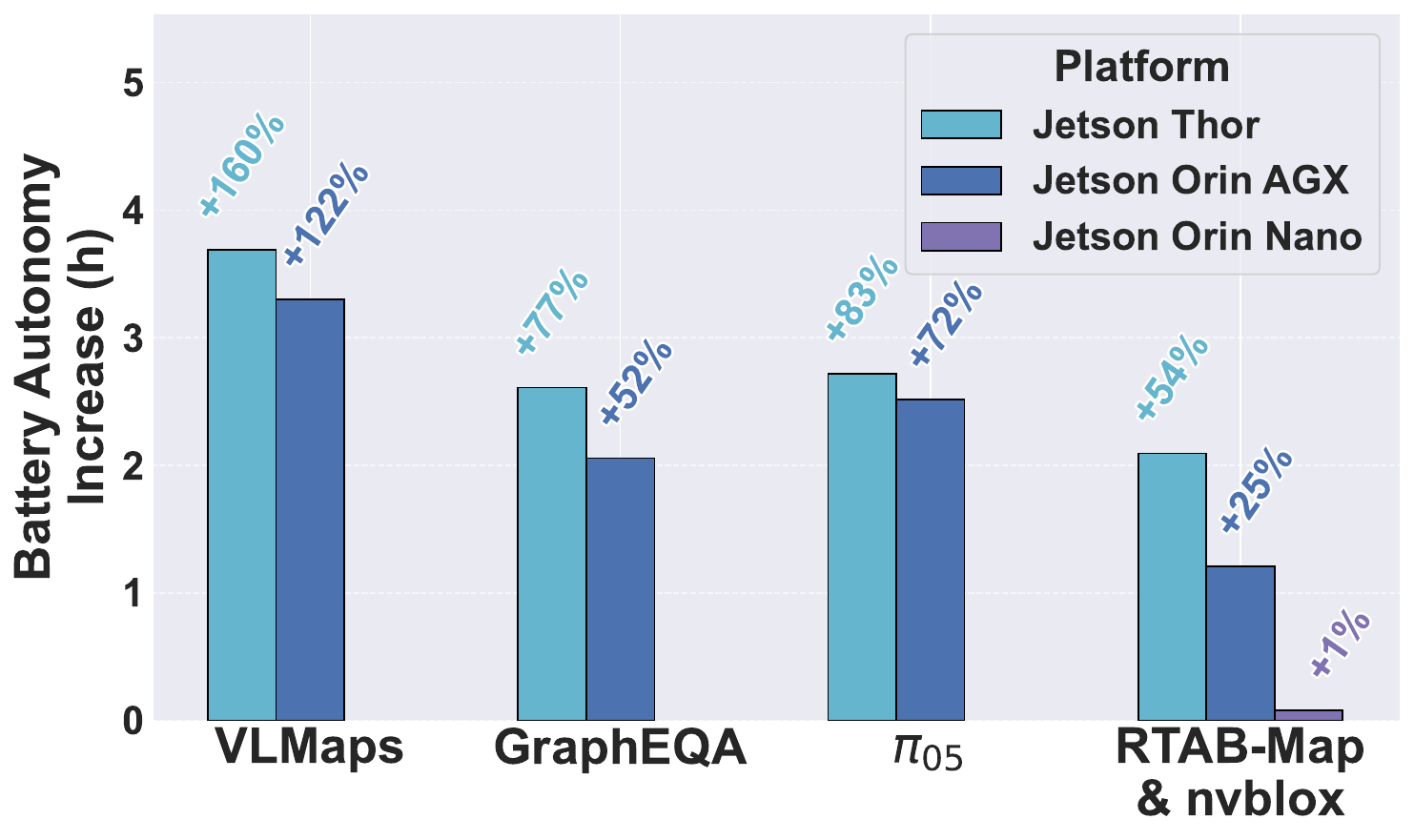}
    \caption{\small
    Stretch 3 battery lifetime increase from offloading, by replacing high-end onboard compute with a Raspberry Pi 5.
    }
    \label{fig:battery_autonomy_increase}
    \vspace{-3mm}
\end{figure}


To put the power numbers into perspective, we estimate the increase in battery lifetime for the Stretch 3 robot if all computations were offloaded to a remote location and the onboard GPUs were replaced with a Raspberry Pi 5 used only for data transmission ($\sim$6W). As shown in Figure~\ref{fig:battery_autonomy_increase}, given the robot's battery capacity of 216Wh and the 30W baseline power consumption of the robot’s sensors and motors, such an offloading strategy could yield substantial battery savings, increasing the robot’s operational lifetime by several hours.
\newline

\noindent\fbox{\parbox{0.97\linewidth}{
\textbf{Takeaway \#2:} \textit{Onboard GPUs significantly increase \\ robot power consumption. Offloading compute can therefore substantially extend a robot's battery lifetime.}
}}

\section{Analysis of Offloading Options}
\label{sec:offloading}



Following the observations of \xref{sec:on_board}, we examine the choices of offloading to a GPU server on the edge and in the cloud.
Two factors determine this choice: (i) end-to-end task latency (\xref{subsec:latency}), and (ii) bandwidth to transmit data to the server (\xref{subsec:bandwidth}).

\subsection{Latency considerations} 
\label{subsec:latency}



\begin{figure}[t]
    \centering
    \includegraphics[width=0.85\columnwidth]{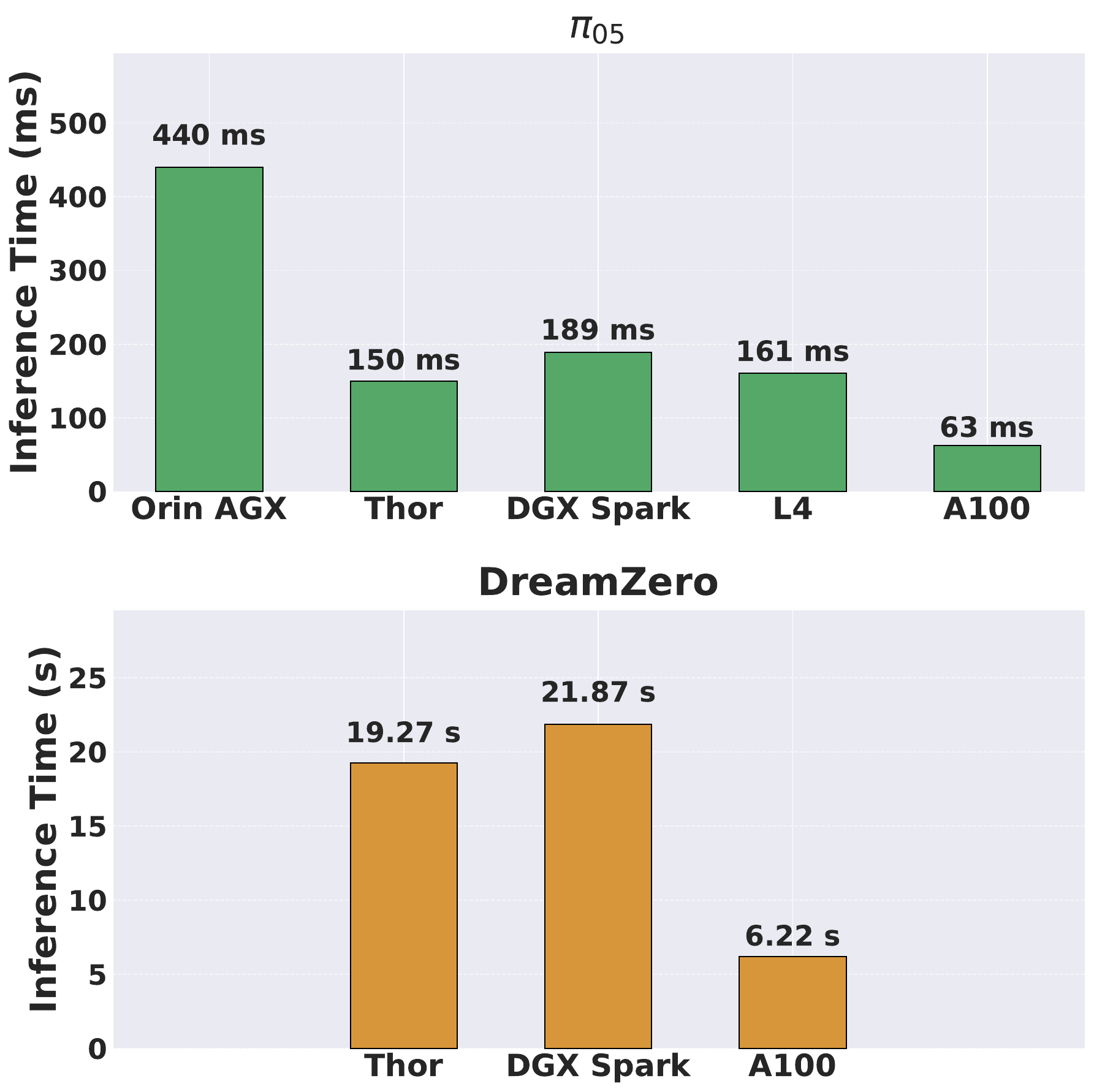}
    \caption{\small 
    Inference time for $\pi_{0.5}$ and DreamZero across different compute platforms. 
    }
    \label{fig:inference_times_vla_wam}
    \vspace{-3mm}
\end{figure}

\begin{figure}[t]
    \centering
    \includegraphics[width=0.9\columnwidth]{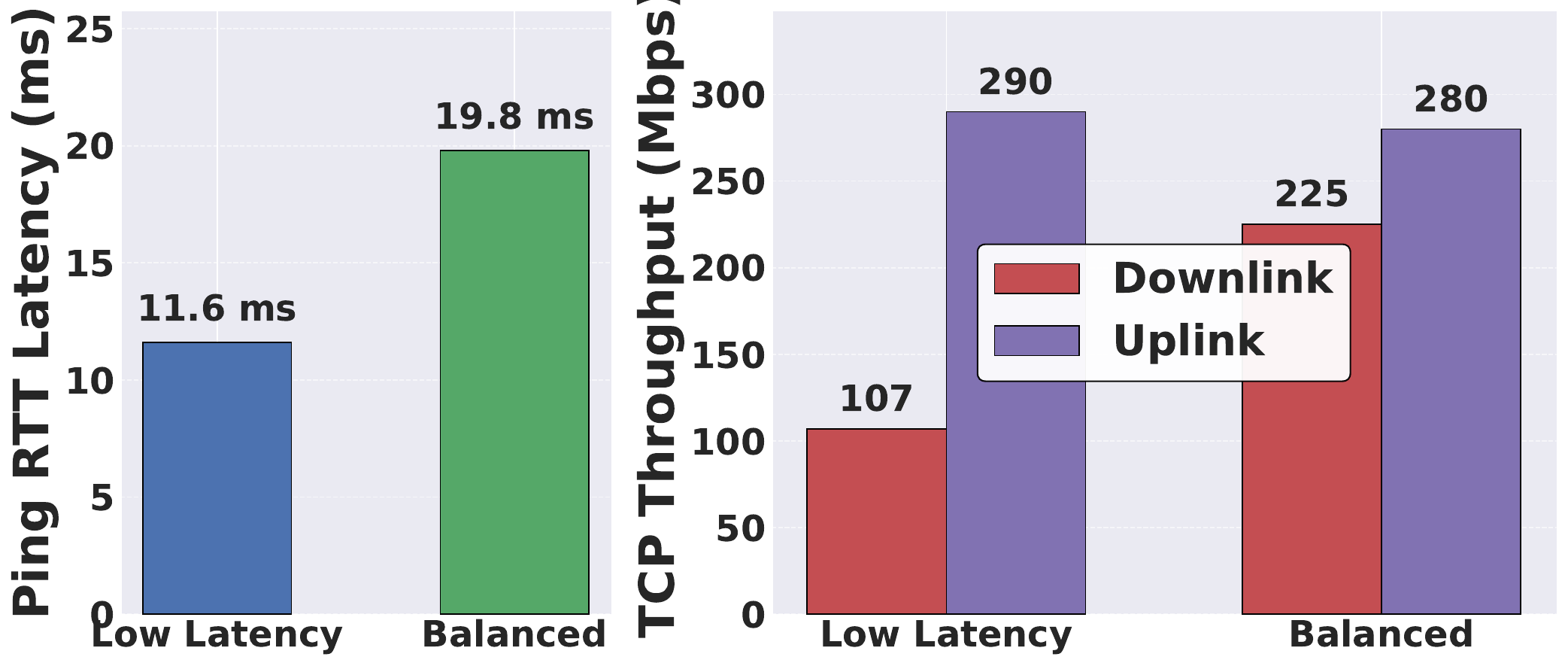}
    \caption{\small 
    Example 5G cell configurations with tradeoffs in terms of latency and throughput, as measured in our testbed.
    }
    \label{fig:5g_cell_tradeoffs}
\end{figure}

\noindent{\bf Latency components:} 
The end-to-end task latency when offloading depends on three factors: the compute latency, latency of the wireless link to reach the base station or edge compute, and the wide-area latency (if the server is in the cloud).
All three latency components can vary depending on the infrastructure characteristics. As shown in Figure~\ref{fig:inference_times_vla_wam}, the compute latency 
of $\pi_{0.5}$ and DreamZero (\xref{subsec:workload-manipulation}) vary with the compute platform: edge platforms, such as the DGX Spark and the L4 server, introduce higher inference latencies compared to server-grade GPUs, like the A100. 
Wired latency 
is affected by the choice of the offloading location and link type~\cite{10.1145/3341302.3342073}. 
Similarly, wireless link latencies 
can be affected by several parameters, such as the wireless technology (e.g., Wi-Fi or 5G), its configurations, and the number of robots sending data.
For example, Wi-Fi latency can be from a few ms in an idle network to tens of ms in a busy one (Figure~\ref{fig:wifi_ping_latency}). 
5G latency additionally depends on the cell configuration (Figure~\ref{fig:5g_cell_tradeoffs}).


\begin{figure}[t]
    \centering
    \includegraphics[width=0.85\columnwidth]{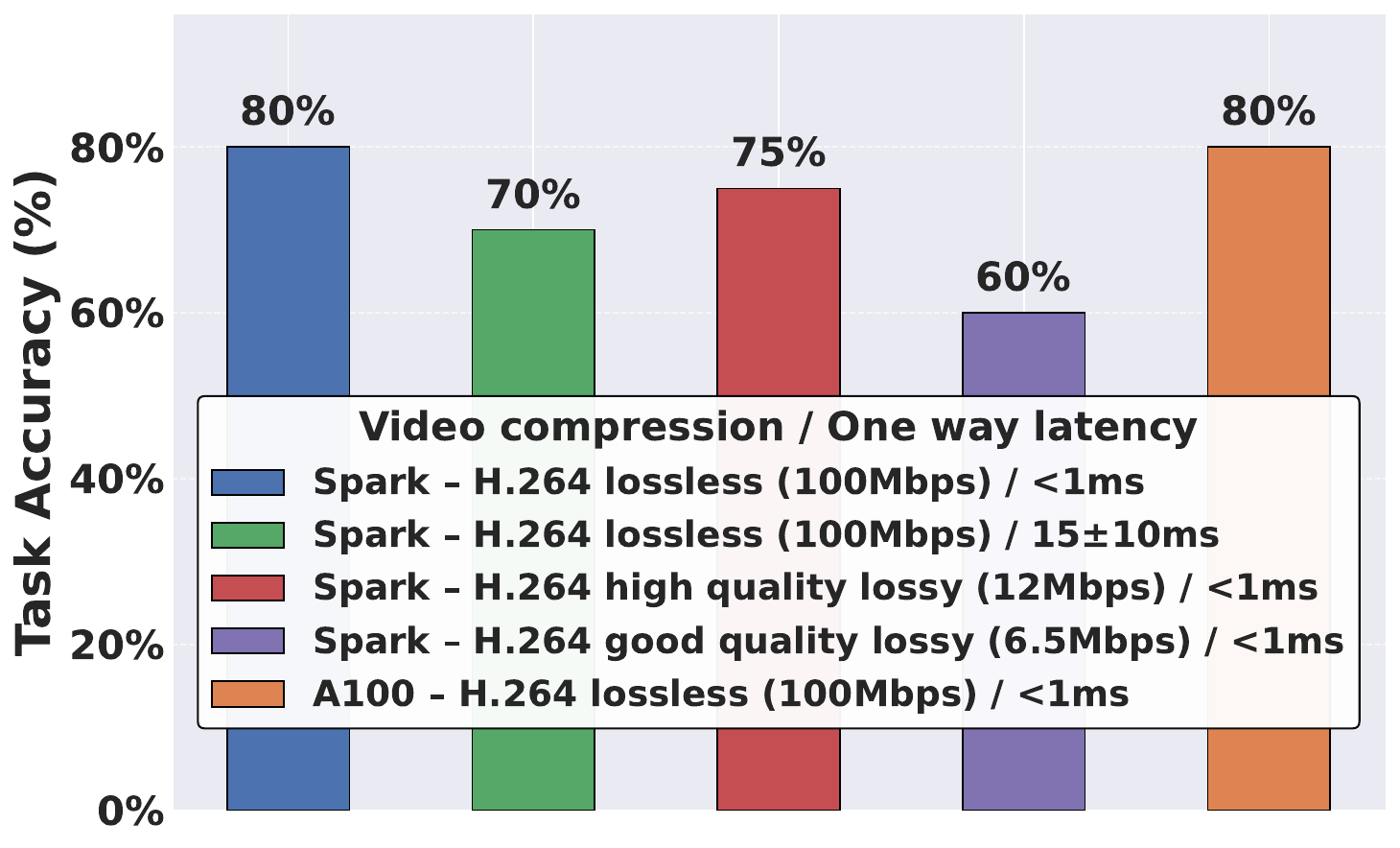}
    \caption{\small 
    $\pi_{0.5}$ accuracy with different latency and bandwidth.
    }
    \label{fig:vla_latency_bandwidth}
    \vspace{-3mm}
\end{figure}

\noindent{\bf Latency vs. Accuracy:} We measure the impact of increased network latency on the accuracy of robotics workloads.
As shown in Figure~\ref{fig:vla_latency_bandwidth}, while the accuracy of the $\pi_{0.5}$ workload is 80\% on the DGX Spark when the network connectivity is ideal, it drops to 70\% once we introduce additional network latency (normal distribution with 10ms mean and 15ms standard deviation one-way~\cite{10.1145/3341302.3342073}). 
The loss of accuracy with added network latency can be compensated by higher-end GPUs.
For example, by replacing the DGX Spark with the A100 VM, accuracy goes back to 80\%, albeit at increased cost. 
We have observed similar trends for RTAB-Map \& nvblox for the same hardware and latency setups, where the accuracy drops to 80\%, compared to the 100\% under ideal network conditions.

\subsection{Bandwidth considerations} 
\label{subsec:bandwidth}



\begin{table}[t]
    \centering
    \footnotesize
    \begin{tabular}{lcccc}
    \toprule
    {\bf Config.} & 15 FPS & 10 FPS & 5 FPS & 1 FPS \\
    \midrule
    640x480 & 20.3 / \hspace{1mm} --\hspace{3mm}  & 13.7 / 91.5\% & 7.2 / 89.8\% & 1.8 / 71.2\% \\
    320x240 & 5.4 / 84.7\% & 3.8 / 84.7\% & 2.1 / 79.7\% & 0.7 / 69.5\% \\
    160x120 & 1.7 / 81.4\% & 1.2 / 79.7\% & 0.8 / 78.0\% & 0.5 / 72.9\% \\
    \bottomrule
    \end{tabular}
    \caption{\small Average bandwidth (Mbps) / recall with VLMaps across image resolutions and frame rates.}
    \label{tab:vlmaps-bandwidth}
\end{table}

\begin{table*}[t]
\centering
\footnotesize
\begin{tabular}{l|ccc|ccc|ccc|ccc|ccc}
\toprule
 & \multicolumn{3}{c|}{\textbf{C1. Batch Runtime (s)}} & \multicolumn{3}{c|}{\textbf{C2. Speedup vs.\ Seq.}} & \multicolumn{3}{c|}{\textbf{C3. Latency vs.\ Thor B=1}} & \multicolumn{3}{c|}{\textbf{C4. GPU Mem (GB)}} & \multicolumn{3}{c}{\textbf{C5. Mem.\ Savings}} \\\hline
Model & B=1 & B=2 & B=4 & B=1 & B=2 & B=4 & B=1 & B=2 & B=4 & B=1 & B=2 & B=4 & B=1 & B=2 & B=4 \\
\midrule
$\pi0.5$ \cite{pi0.5}        & 0.0633 & 0.0956 & 0.158 & 1.00 & 1.32 & 1.60 & +56.1\% & +33.7\% & -9.5\% & 7.90 & 8.25 & 8.70 & 0\% & 47.8\% & 45.6\% \\\hline
SAM \cite{sam}          & 0.366 & 0.725 & 1.448 & 1.00 & 1.01 & 1.01 & +12.9\% & -45.5\% & -79.3\% & 2.52 & 2.53 & 2.56 & 0\% & 49.3\% & 71.5\% \\\hline
CLIP \cite{clip}            & 0.0758 & 0.1398 & 0.263 & 1.00 & 1.08 & 1.16 & -10.6\% & -8.7\% & -31.7\% & 3.90 & 3.96 & 4.08 & 0\% & 49.7\% & 47.9\% \\\hline
Depth &  &  & &  &  &  &  &  &  &  &  &  &  &  &  \\
Anything \cite{depthanything}  & 0.0722 & 0.133 & 0.273 & 1.00 & 1.08 & 1.06 & +4.8\% & -14.5\% & -27.2\% & 1.28 & 1.29 & 1.29 & 0\% & 49.7\% & 74.8\% \\\hline
Qwen2.5 \cite{qwen2.5}   & 1.217 & 1.262 & 1.368 & 1.00 & 1.93 & 3.55 & +27.2\% & +24.4\% & +18.1\% & 7.52 & 7.77 & 8.25 & 0\% & 48.4\% & 72.6\% \\
\bottomrule
\end{tabular}
\caption{\small Batching of robotic models on A100 GPU. (C1.) \textbf{Batch Runtime} is the time to process all $B$ requests as a batch. (C2.) \textbf{Speedup}  relative to running $B$ requests one after another. (C3.) \textbf{Latency vs.\ Thor} is the latency relative to a single Thor request ($+$ = faster, $-$ = slower). (C4.) Memory utilizations. (C5.) \textbf{Mem.\ Savings vs.\ $B$ Inst.} is the memory reduction compared to loading $B$ separate instances.}
\label{tab:batch_benchmark}
\end{table*}

When offloading workloads, the robot must continuously transmit sensor data over the network. The feasibility of such offloading therefore depends not only on latency, but also on sustained network bandwidth.
We explore the impact of bandwidth using $\pi_{0.5}$ and VLMaps as representative examples. 

In the case of $\pi_{0.5}$, we stream lossless video from three cameras mounted on the bimanual SO-101 setup with a resolution of $640\times480$ at 30 FPS.
This requires an average uplink bandwidth of $\sim$100Mbps and results in a task accuracy of 80\%, as shown in Figure~\ref{fig:vla_latency_bandwidth}.
In multi-robot settings, bandwidth demands scale with the number of robots, which can quickly saturate the network capacity, even with tens of robots. 

\noindent{\bf Bandwidth vs. Accuracy:} To alleviate the stress on the network bandwidth, we experiment with lossy H.264 video compression with two different quality configurations, ``high'' and ``good'', which result in bandwidth requirements of 12 and 6.5 Mbps, respectively.
We ensure that no packets are lost, which could lead to corrupt frames.
As we can observe in Figure~\ref{fig:vla_latency_bandwidth}, the bandwidth reduction comes at the cost of a drop in the accuracy of the handover and pick-and-place task.

Similar observations can be made about the performance and bandwidth tradeoffs of VLMaps. 
Here, we reduce the bandwidth for streaming the robot's RGB images by reducing the frame rate and resolution of the video (Table~\ref{tab:vlmaps-bandwidth}).
We treat the map generated by VLMaps with the $640\times480$, 15 FPS configuration as a reference, which identifies 59 unique objects by matching map embeddings against a large set of indoor text labels. For the alternative resolutions and frame rates, we compute recall as the percentage of these 59 objects that are also identified (Table~\ref{tab:vlmaps-bandwidth}). 
Reducing frame rate has a considerable effect: at $640\times480$, recall drops to 91.5\% at 10 FPS and 71.2\% at 1 FPS. Resolution reductions similarly degrade recall. At 15 FPS, lowering resolution from $640\times480$ to $320\times240$ yields 84.7\% recall, and further reducing to $160\times120$ yields 81.4\%.
\newline

\noindent\fbox{\parbox{0.97\linewidth}{
\noindent\textbf{Takeaway \#3:} \textit{Obtaining the potential benefits of offloading robotics workloads requires carefully considering the complex tradeoff involving workload performance, network latency, bandwidth, and available GPU resources.}
}}

\section{Multiplexing Robotic Workloads}
\label{sec:multiplexing}

Offloading unlocks opportunities for multiplexing robotic workloads for reducing the deployment cost.
In this section, we explore the opportunities and challenges to such sharing.

\subsection{Sharing opportunities}
\label{sec:sharing_opportunities}

\subsubsection{\bf Batching:}
To understand the extent to which sharing of edge/cloud compute resources is possible, we first consider several robots simultaneously performing the same type of task and explore the extent to which robotic workloads can benefit from batching requests to the corresponding models.
We focus on VLMaps, GraphEQA, and $\pi_{0.5}$, given that batching does not apply to nvblox.
In the case of VLMaps, we decompose the process to its core underlying models (i.e., CLIP, SAM and Depth Anything v2).
GraphEQA utilizes Qwen-VL-2.5-3B, which we study using generation responses of 32 tokens.
Table~\ref{tab:batch_benchmark} shows the results for the A100 GPU for batches (B = 1, 2, 4). As we can observe (column C5), batching brings significant memory savings for all models, ranging from 45.6\% for $\pi_{0.5}$ to 74.8\% for DepthAnything (when B=4).

Inference latency improves significantly compared to sequential execution of the inference requests (column C2), with $\pi_{0.5}$ and Qwen models speeding up by 1.6 and 3.55, respectively.
As we see in column C3, by serving all four requests in parallel on the more powerful A100 GPU, inference latency increases only marginally for $\pi_{0.5}$ while decreasing significantly for Qwen, relative to a single request on the onboard Thor GPU. 
This means the A100 can serve 4 robots simultaneously with  latency comparable to {\em each} robot having its own dedicated Thor.
In the case of SAM, CLIP, and Depth Anything, the inference latency increases roughly linearly with the batch size. The vision models (SAM, CLIP, and Depth Anything) do not benefit from sharing the prefill phase as in the case of LLM-based workloads ($\pi_{0.5}$ and Qwen) ~\cite{patel2024splitwise, jiang2026fast}.
\newline


\noindent\fbox{\parbox{0.97\linewidth}{
\noindent\textbf{Takeaway \#4:} \textit{Robotic workloads involving VLAs and VLMs present significant opportunities for sharing GPUs through real-time batching, more so than pure vision models that are typically used in perception tasks.}
}}



\subsubsection{\bf Statistical multiplexing opportunities:}
\begin{figure}[t]
    \centering
    \includegraphics[width=0.85\columnwidth]{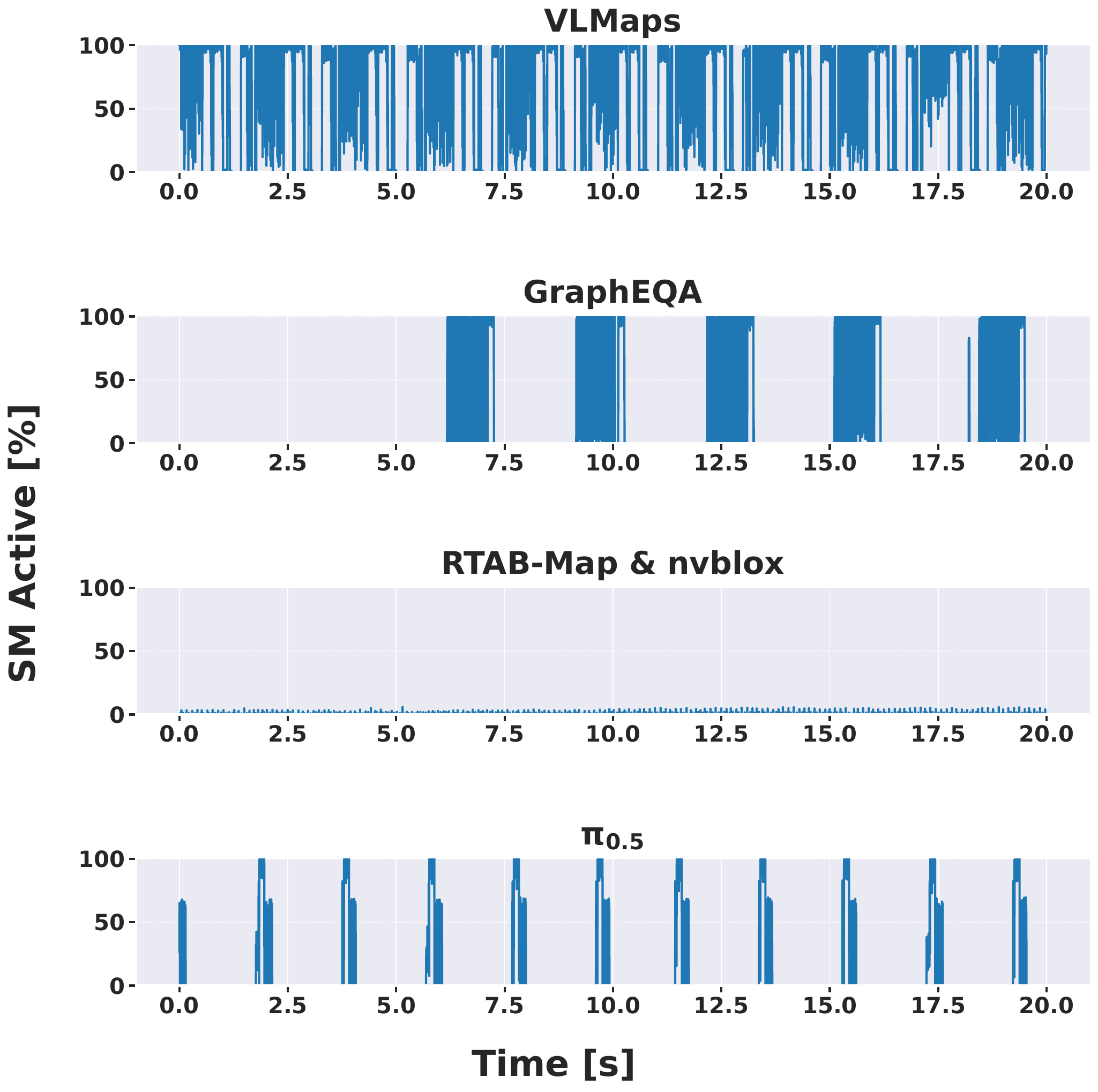}
    \caption{\small
    SM utilization for studied workloads on DGX Spark.
    }
    \label{fig:statistical_muxing_spark}
\end{figure}

\begin{figure}[t]
    \centering
    \includegraphics[width=0.85\columnwidth]{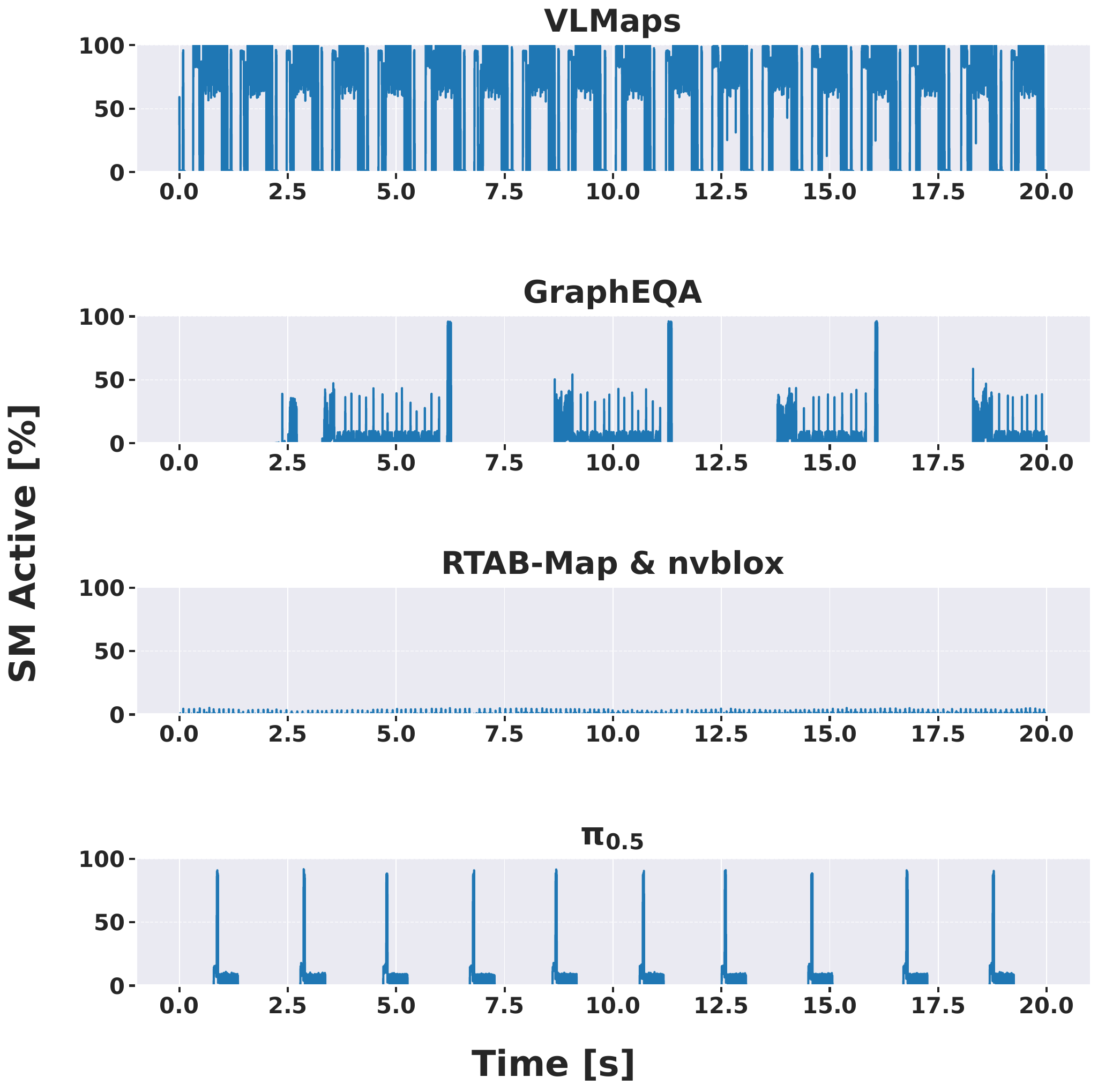}
    \caption{\small
    SM utilization for studied workloads on A100 VM.
    }
    \label{fig:statistical_muxing_a100}
    \vspace{-3mm}
\end{figure}

We use the Nvidia \verb|nsys| profiling tool to measure the GPU Streaming Multiprocessors (SM) utilization for the various workloads under study.
As illustrated in Figures~\ref{fig:statistical_muxing_spark}~and~\ref{fig:statistical_muxing_a100} for the DGX Spark and the A100 VM, all the workloads present patterns of periodic activity, followed by idle GPU periods, exposing ample opportunities for the multiplexing of workloads.
The periodicity of active and idle periods depends on the task, with the idle gaps being larger in the case of the A100 GPU.
In the case of the $\pi_{0.5}$ model and GraphEQA, the active GPU periods correspond to the generation of actions for the robot to execute (move the arms for $\pi_{0.5}$ and explore a new area for GraphEQA), while the idle periods correspond to the time of executing those actions or sitting idle.
On the other hand, VLMaps and RTAB-Map \& nvblox are constantly processing frames, with idle GPU periods corresponding to CPU activity (e.g., to transfer frames to and from the GPU).

All the workloads exhibit low absolute utilization or reach their peaks only briefly (e.g., the peaks of GraphEQA and $\pi_{0.5}$ only last for a few tens of milliseconds).
This means that significant sharing opportunities exist even during active periods.
It should be noted that while VLMaps is more computationally demanding in the average case, its best-effort nature allows the delay of its execution, to prioritize other more latency-sensitive workloads, such as $\pi_{0.5}$.
Furthermore, the semantic map indexing of VLMaps could also be paused while the robot is not moving around, since no new information is added from steady frames, freeing up even more compute resources.
\newline

\noindent\fbox{\parbox{0.97\linewidth}{
\noindent\textbf{Takeaway \#5:} \textit{Robotic workloads have alternating idle and busy periods that present opportunities for statistical multiplexing, which become more ample with higher-grade GPUs.}
}}

\begin{figure}[t]
    \centering
    \includegraphics[width=\columnwidth]{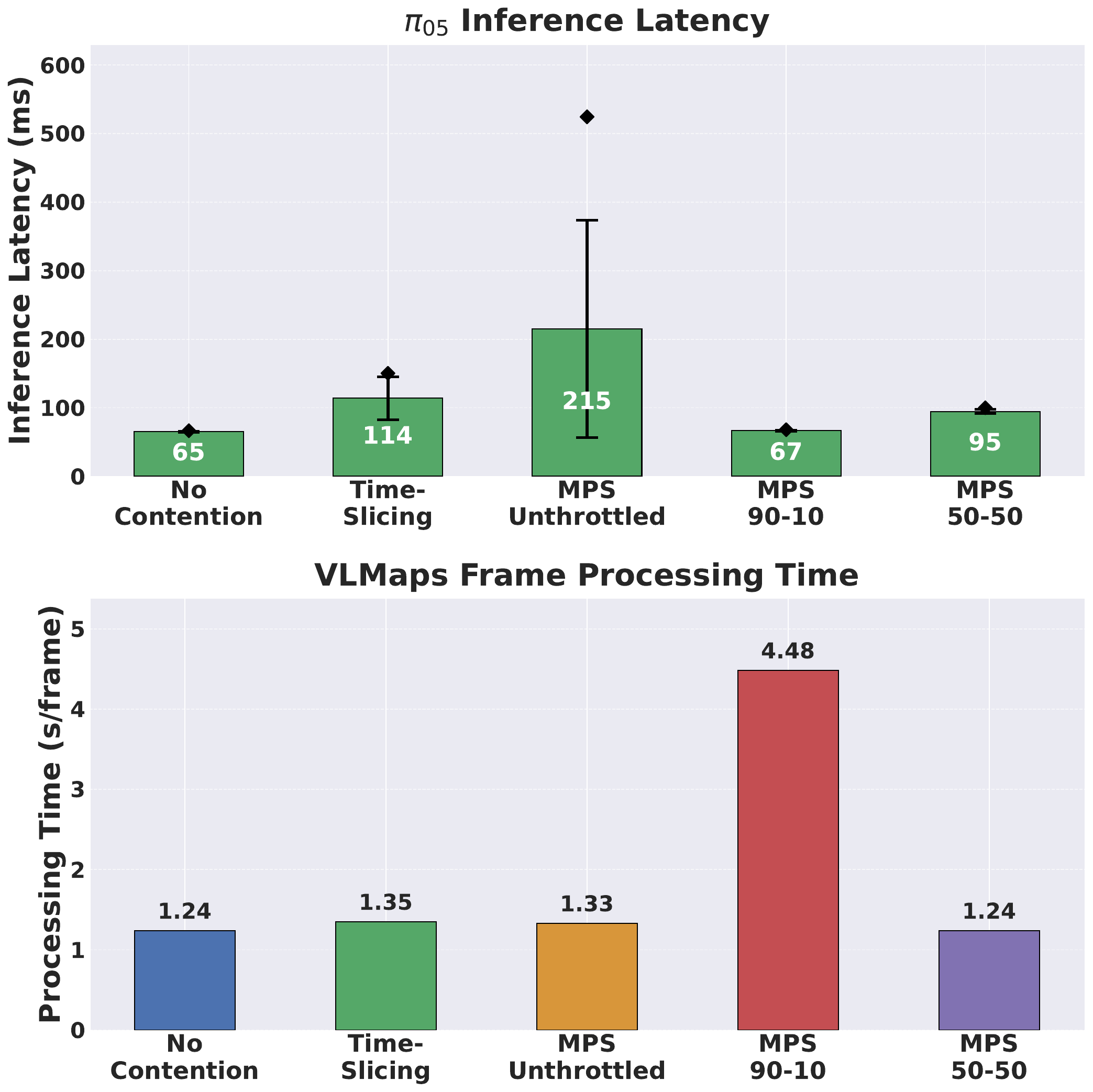}
    \caption{\small
    Impact of $\pi_{0.5}$ and VLMaps sharing the compute.
    }
    \label{fig:compute_interference}
\end{figure}

\subsection{Sharing challenges}

\subsubsection{\bf Compute sharing:}
Here we focus on the performance implications of statistical multiplexing.
We present one concrete example of sharing the A100 VM between $\pi_{0.5}$ and VLMaps, but note that our findings extend to the other workloads under study.
For this experiment, we separate the two workloads to run on different CPU sets, using \verb|taskset|, and consider four different widely-used policies for sharing the GPU: i) default GPU scheduling with time-slicing, ii) MPS scheduling, where both workloads are not throttled by the GPU scheduler, iii) MPS scheduling, where the latency critical $\pi_{0.5}$ workload is given 90\% of the GPU SMs, leaving 10\% for VLMaps, and iv) MPS scheduling with 50-50 assignment of SMs between the two workloads. In all four cases, we measure the inference latency of $\pi_{0.5}$, as well as the time required to process a single frame with VLMaps.

In Figure~\ref{fig:compute_interference}, we observe that sharing the GPU using time slicing (i) and  unthrottled MPS (ii) results in an inference latency increase of 75\% and 230\%, respectively.
This is because, as shown in Figures~\ref{fig:statistical_muxing_spark}~and~\ref{fig:statistical_muxing_a100}, the VLMaps workload is more computationally intensive, resulting in long scheduling delays for $\pi_{0.5}$.
By introducing throttles, and guaranteeing higher SM availability for $\pi_{0.5}$, its inference latency reduces.
In the optimal case, when restricting VLMaps to 10\% of the SMs (iii), the latency increases by just 3\%.
However, this static throttling results in a significant increase of 261\% in the frame processing time of VLMaps, since the SMs allocated to $\pi_{0.5}$ can no longer be reallocated to VLMaps even during idle periods. 



\subsubsection{\bf Wireless spectrum sharing:}
As with the scheduling of compute, high latency can be a critical issue when sharing radio resources for offloading the workloads of several robots.
To demonstrate, we capture a trace of the traffic that our Stretch 3 robot sends to the offloading server for the GraphEQA model ($\sim$50 Mbps uplink traffic), and we replay this traffic over our Wi-Fi 6 network for a number of robot clients, while measuring the ping RTT of another (idle) client.

\begin{figure}[t]
    \centering
    \includegraphics[width=0.7\columnwidth]{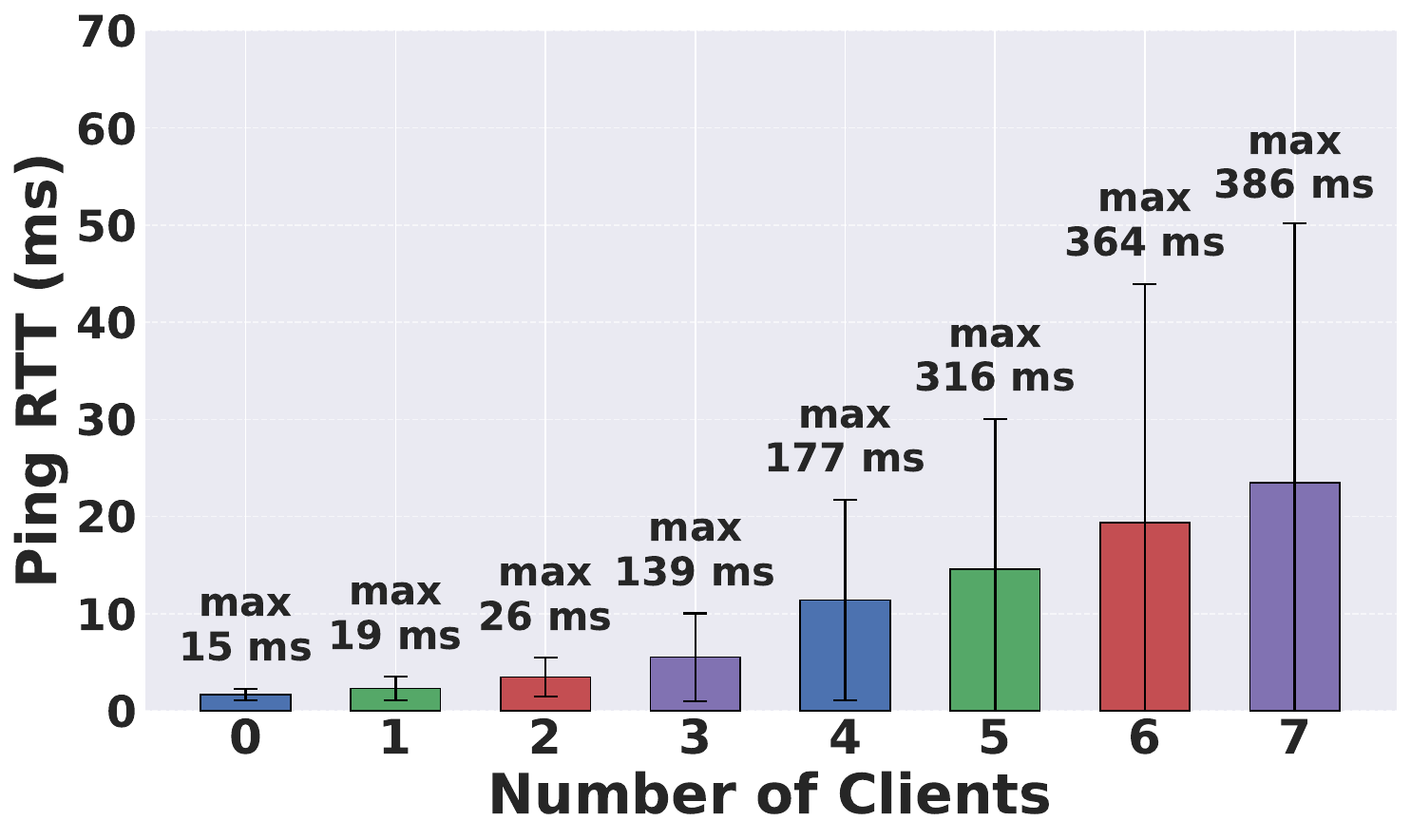}
    \caption{\small
    Ping RTT of idle client with varying additional connected clients, each generating 50Mbps of UL TCP traffic.
    }
    \label{fig:wifi_ping_latency}
    \vspace{-3mm}
\end{figure}

As Figure~\ref{fig:wifi_ping_latency} shows, the RTT of the idle client significantly increases with the number of active clients, and can reach close to 25ms in the mean and up to 50ms in the standard deviation.
We note that this latency increase was observed, despite the several optimizations in our Wi-Fi network to ensure low latency, including using OFDMA and centralized scheduling, as well as ensuring no interference from neighboring access points.
As with unthrottled compute sharing, the scheduling delays of Wi-Fi increases with the contending users ~\cite{foukas2025future, ghosh2026beyond}.
\newline


\noindent\fbox{\parbox{0.97\linewidth}{
\noindent\textbf{Takeaway \#6:}
\textit{In sharing compute and wireless networks between robots, it is challenging to balance real-time and high-throughput workloads, necessitating dynamic QoS-aware scheduling, traffic shaping, and network slicing.}
}}

\section{Need for a Distributed Robotics Platform}
\label{sec:discussion}

Our measurement study highlights the growing need for powerful GPUs to support emerging mobile robotic manipulation workloads, as well as the potential for cost-effective shared infrastructure. Furthermore, we demonstrate that no single deployment strategy--onboard, edge, or cloud--is optimal for all deployment scenarios and use cases due to inherent tradeoffs across workload performance, network latency, bandwidth, power consumption, and monetary cost.

These characteristics motivate the need for a {\em distributed robotics inference platform} that spans robots, edge, and cloud resources. Such a platform should support distributed orchestration of workloads based on bandwidth, latency, and compute requirements, and enable dynamic, QoS-aware scheduling of GPU, CPU, and network resources.
Although similar ideas have been explored in domains such as mobile~\cite{maui} and telecommunications~\cite{ananthanarayanan2025distributed}, robotic systems introduce new constraints. In particular, closed perception–action loops impose strict real-time requirements, creating new challenges for system design and implementation to ensure safety and reliable performance in physical-world tasks.
\section{Related Work}

While foundation models have significantly advanced the state of mobile robotic manipulation, most existing research emphasizes algorithmic accuracy and generalization, often overlooking the system-level constraints. 

\noindent{\bf 1) System constraints:}
Current mobile robotic manipulation literature primarily evaluates task performance under idealized or simulated conditions \cite{ovmm}. While these works demonstrate impressive capabilities, they often assume that substantial compute resources are readily available. As a result, the impact of heterogeneous onboard compute platforms remains under-explored.
In parallel, the robotics community has studied energy-efficient motion planning and control \cite{energy}. However, little work has examined how modern AI workloads affect the energy consumption and battery longevity of robots.
Our work addresses this gap by systematically measuring the energy, memory, and performance trade-offs of core robotics workloads across several onboard compute platforms (\xref{sec:on_board}). To our knowledge, this is the first study to quantify how task accuracy and execution time of robotics foundation models vary as a function of system-level resource constraints.

\noindent{\bf 2) Offloading considerations:} 
A natural strategy for handling resource-intensive workloads is offloading to edge or cloud devices. Offloading has been extensively studied in mobile and edge computing \cite{maui}, particularly for deep neural network inference on resource-constrained devices \cite{9985008} and large-scale video analytics \cite{10973078}. 
However, such systems generally operate in soft real-time or open-loop settings, where occasional latency fluctuations are tolerable. In contrast, robotic systems operate within closed perception–action loops, where delays or dropped computations can directly degrade control performance or compromise safety. Thus, offloading considerations in robotic systems differ from those in mobile computing. 

Within robotics, offloading has received comparatively limited empirical attention. Early surveys highlighted the potential for cloud infrastructure to accelerate workloads like SLAM and motion planning while noting risks introduced by network latency and connectivity variability \cite{7006734}. More recent studies have quantified these effects, measuring how fixed communication latencies influence trajectory tracking and gesture recognition \cite{10.3389/frobt.2023.1168694}. 
Despite this work, there is little measurement-based understanding of how the modern robotics workloads behave under offloading regimes. Prior studies largely focused on classical robotics pipelines or relatively small neural networks, which do not capture the memory footprint and bandwidth demands of contemporary multimodal models. 
To address this gap, we provide a quantitative analysis of offloading constraints for robotics workloads (\xref{sec:offloading}).
\looseness=-1


\noindent{\bf 3) Resource sharing in multi-robot settings:}
While multi-robot systems are well studied in the context of coordinated control \cite{multi_control} and collaborative mapping \cite{multi_map}, the systems implications of shared compute infrastructure remain largely unexplored.
This paper provides the first comprehensive measurement study of these multiplexing effects for robotics workloads, characterizing both the opportunities and limitations of shared infrastructure in multi-robot deployments (\xref{sec:multiplexing}).

\vspace{-2mm}

\section{Conclusions}

In this work, we performed the first measurement study of mobile robotic manipulation workloads across onboard, edge, and cloud GPU platforms.
Our findings reveal that offloading such workloads can be beneficial in terms of performance and battery savings for the robots, but introduces its own challenges, due to the effects of network latency and bandwidth on task accuracy.
Finally, the study highlights the opportunities and challenges of sharing compute across robot fleets.
We hope that this study will help guide future research in the space of inference systems for mobile robots.
\bibliographystyle{ACM-Reference-Format}
\bibliography{ref}


\end{document}